\documentclass[10pt,twocolumn,letterpaper]{article}

\usepackage{3dv}
\usepackage{times}
\usepackage{epsfig}
\usepackage{graphicx}
\usepackage{amsmath}
\usepackage{amssymb}

\usepackage[numbers]{natbib}

\usepackage[pagebackref=true,breaklinks=true,letterpaper=true,colorlinks,bookmarks=false]{hyperref}

\threedvfinalcopy % *** Uncomment this line for the final submission

\def\threedvPaperID{28} % *** Enter the 3DV Paper ID here

\ifthreedvfinal\fi

\renewcommand{\baselinestretch}{0.938}

\begin{document}

%\title{High Resolution Surface Tracking of Deformable Articulated Objects}
\title{Dynamic High Resolution\\
Deformable Articulated Tracking}

% You will get a Paper-ID when submitting a pdf file to the conference system
%\author{
%Aaron Walsman \\
%{\tt\small awalsman@} \and
%Weilin Wan\\
%{\tt\small weiliw@uw.edu}
%\and Tanner Schmidt\\
%{\tt\small tws10@}
%\and Dieter Fox\\
%{\tt\small fox@} \\
%Paul G. Allen School of Computer Science and Engineering, University of Washington}

%\author{
%Aaron Walsman \and
%Weilin Wan \and
%Tanner Schmidt \and
%Dieter Fox\\
%Paul G. Allen School of Computer Science and Engineering, University of Washington}

%\author{
%Aaron Walsman \\ {\tt\small awalsman@\{1\}} \and
%Weilin Wan \\ {\tt\small weiliw@uw.edu} \and
%Tanner Schmidt \\ {\tt\small tws10@\{1\}} \and
%Dieter Fox \\ {\tt\small fox@\{1\}}
%}

\author{
\qquad Aaron Walsman \hfill
Weilin Wan \hfill
Tanner Schmidt \quad \hfill
Dieter Fox \ \quad \phantom{M} \\
{\tt\footnotesize awalsman@cs.washington.edu} \qquad
{\tt\footnotesize weiliw@uw.edu} \qquad
{\tt\footnotesize tws10@cs.washington.edu} \qquad
{\tt\footnotesize fox@cs.washington.edu} \vspace{3mm} \\
%{\tt\small \{awalsman, tws10, fox\}@cs.washington.edu} \qquad
%{\tt\small weiliw@uw.edu}\\
%{\tt\small awalsman@\{1\}} \hfill
%{\tt\small weiliw@\{2\}} \hfill
%{\tt\small tws10@\{1\}} \hfill
%{\tt\small fox@\{1\}}\\
Paul G. Allen School of Computer Science and Engineering University of Washington
}

%\author{
%Aaron Walsman, Weilin Wan, Tanner Schmidt, Dieter Fox \\
%{\tt\small awalsman@cs.washington.edu, weiliw@uw.edu,
%tws10@cs.washington.edu, fox@cs.washington.edu} \\
%Paul G. Allen School of Computer Science and Engineering, University of Washington}

%\and

% avoiding spaces at the end of the author lines is not a problem with
% conference papers because we don't use \thanks or \IEEEmembership

%Homer Simpson\authorrefmark{2},
%Montgomery Scott\authorrefmark{3} and
%Eldon Tyrell\authorrefmark{4}}
%Telephone: (800) 555--1212, Fax: (888) 555--1212}
%\authorblockA{\authorrefmark{4}Tyrell Inc., 123 Replicant Street, Los Angeles, California 90210--4321}}

\maketitle

%%%%%%%%%%%%%%%%%%%%%%%%%%%%%%%%%%%%%%%%%%%%%%%%%%%%%%%%%%%%%%%%%%
%%%%%%%%%%%%%%%%%%%%%%%%%%%%%%%%%%%%%%%%%%%%%%%%%%%%%%%%%%%%%%%%%%
\begin{abstract}
%%%%%%%%%%%%%%%%%%%%%%%%%%%%%%%%%%%%%%%%%%%%%%%%%%%%%%%%%%%%%%%%%%
%%%%%%%%%%%%%%%%%%%%%%%%%%%%%%%%%%%%%%%%%%%%%%%%%%%%%%%%%%%%%%%%%%
%We present a 3D template-based tracking system for articulated
%deformable objects.  Our system is able to track human body pose
%and high resolution surface contours in real time using
%a commodity depth sensor and GPU hardware.
%We implement this as two interleaved optimization processes:
%one that updates a skeleton to account for changes in pose,
%and another that updates the vertices of a high resolution mesh to
%track the subject's shape.  Using this approach we are able to
%capture dynamic sub-centimeter surface detail such as folds
%and wrinkles in clothing that are typically only possible in
%template-free reconstruction techniques.  However in contrast to
%these reconstruction techniques, our template-based approach
%provides useful semantic consistency across different tracking
%sessions and different subjects.  The end result
%is highly accurate spatiotemporal and semantic information which
%is necessary for physical human robot interaction.
%Our system has additional applications in interactive
%systems, entertainment and augmented reality.
The last several years have seen significant progress
in using depth cameras for tracking articulated objects
such as human bodies, hands, and robotic manipulators.  Most
approaches focus on tracking skeletal parameters of a fixed
shape model, which makes them insufficient for
applications that require accurate estimates of deformable
object surfaces.  To overcome this limitation, we present a 3D
model-based tracking system for articulated deformable
objects.  Our system is able to track human body pose
and high resolution surface contours in real time using
a commodity depth sensor and GPU hardware.  We implement
this as a joint optimization over a skeleton to account
for changes in pose, and over the vertices of a high
resolution mesh to track the subject's shape.
Through experimental results we show that we are
able to capture dynamic sub-centimeter surface detail
such as folds and wrinkles in clothing.  We also show
that this shape estimation aids kinematic pose
estimation by providing a more accurate target to match
against the point cloud.
The end result is highly accurate spatiotemporal and
semantic information which is well suited for
physical human robot interaction as well as virtual
and augmented reality systems.
\end{abstract}

%Our approach provides useful semantic consistency across different tracking sessions and different subjects.

%\IEEEpeerreviewmaketitle

\section{Introduction}
% Paragraph one: overview
%In order for robots to interact with complex
%deformable objects, a vision system must produce a perceptual
%representation that combines both spatial and semantic information,
%and must be fast enough to keep up with the object's physical motion.
%This is especially true of systems that interact
%directly with humans where safety is critical and
%mistakes can be dangerous.
% Paragraph two: primary contribution
We present a real-time tracking system capable of estimating
the surface and kinematic pose of deformable objects
using model-based optimization.
Our surface estimation not only adapts to match a particular
subject, but does so dynamically, tracking complex surface
details such as cloth folds and wrinkles as they appear and
disappear.  Through experiments we show that this tracker is
capable of simultaneously capturing human body pose and
sub-centimeter surface detail in real time.
%Our method takes
%inspiration from recent techniques in articulated tracking and
%surface reconstruction.

Our work has applications in virtual and augmented reality systems
that require real time human reconstruction for telepresence,
performance capture and games.  There are also many practical
applications in robotics systems that require precise spatial
information about the surface of humans and deformable objects.
%Our work has applications in robotics systems that require
%precise spatial information about the surface of humans and
%deformable objects, as well as virtual and augmented reality
%applications that require real time human reconstruction
%for telepresence, performance capture and games.
For example there are several applications in robotic personal
assistance, health care and rehabilitation that are critically
hampered by a lack of reliable human pose and surface estimation.

%There are two motivating applications behind our work.
%The first is robotics problems that require precise
%spatial information about the surface of humans and
%deformable objects. There are several robotics
%applications in personal assistance, health care and
%rehabilitation that are critically hampered by a lack
%of reliable human pose and surface estimation.
%The second motivation is virtual and augmented
%reality applications that require real time human
%reconstruction for telepresence, performance capture
%and games.

%%%%%%%%%%%%%%%%%%%%%%%%%%%%%%%%%%%%%%%%%%%%%%%%%%%%%%%%%%5
\begin{figure}[t]
\centering
\includegraphics[width=0.45\textwidth]{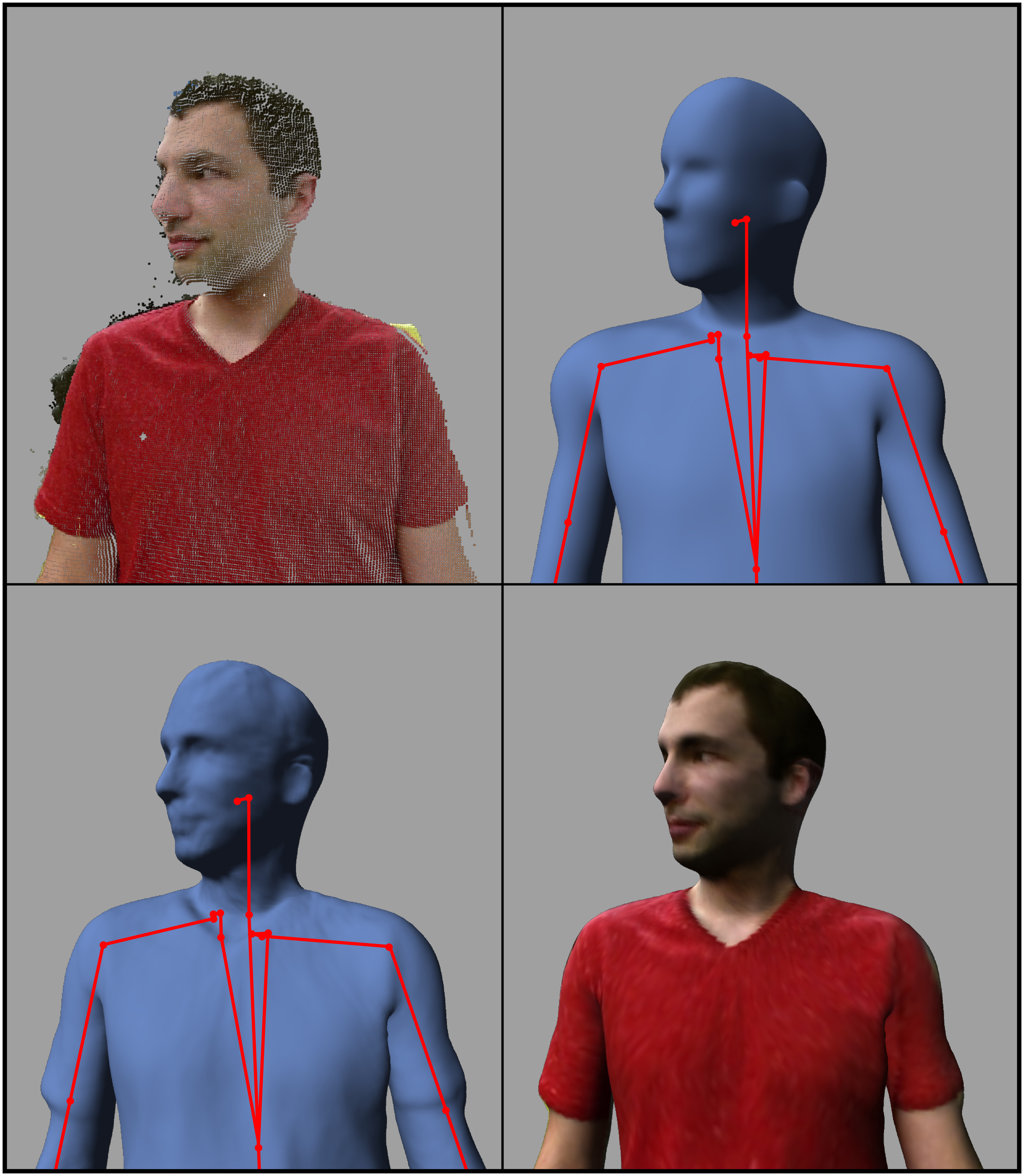}
\caption{Our model tracking a point cloud.
Top left: Colored point cloud input.
Top Right: Estimated skeleton and surface mesh without
surface tracking.  Bottom Left: High resolution mesh tracking
the dynamic shape of the subject.  Bottom Right: The high
resolution mesh with projected colors.}
\label{fig:one}
\end{figure}
%%%%%%%%%%%%%%%%%%%%%%%%%%%%%%%%%%%%%%%%%%%%%%%%%%%%%%%%%%%

% Paragraph three: how does it work?  Broad strokes.
The goal in creating this system is to combine recent
advances in dynamic surface reconstruction with fast articulated
tracking techniques.
Our approach fits a skeletal model and high-resolution
polygon mesh to a point cloud.  The skeleton is designed to
capture the underlying kinematic structure of the subject and
estimate it's large-scale motion, while the polygon mesh captures
volume differences between subjects and more complex surface
details.
This produces both a low-dimensional
pose that can be used for gesture and activity
recognition, as well as a dense estimation of the surface
which can be used for precise physical interaction.
Furthermore, because our mesh comes from a predefined
template model, it is semantically consistent across
capture sessions with different subjects.  This means
that we can determine not only where the surface of the
subject is, but which locations on this surface correspond
to specific regions and body parts.

%Besides the useful information that each of these components
%provide, they also form a mutually beneficial relationship
%within this
%system.  The skeleton optimization uses the current estimate of
%the shape when computing pose updates, so any improvements to
%the mesh will improve the pose estimation.  At the same time
%the polygon mesh follows the skeleton so new pose estimates give
%the shape optimization a better initialization.

%The primary motivations behind our work are robotics
%applications that require precise spatial information about
%humans and deformable objects.  There are several robotics
%applications in personal assistance, health care and
%rehabilitation that are hampered by a lack of
%reliable human pose and surface estimation.
%Figure \ref{fig:shave} is from the work of
%Grice, Hawkins et al.
%\cite{grice2013whole, hawkins2014assistive} and
%shows one application in which a robot
%is helping a person with limited arm mobility shave.
%This level of physical interaction can only be accomplished
%safely when it is paired with accurate surface information.
%In addition to robotics, our work has
%further applications in augmented reality,
%performance capture and interactive games.

This paper is organized as follows.
Section \ref{sec:related_work} below discusses related work and
our relationship to existing approaches.  Section
\ref{sec:method_template_model} describes the details of our
model and then Section \ref{sec:method_optimization} explains
how we fit this model to observations.  We then show experimental
results in Section \ref{sec:experiments} and conclude in
Section \ref{sec:conclusion}.

%Each component of this estimation is simultaneously beneficial to
%the other component and a useful output in it's own right.
%The skeleton provides low dimensional pose information that can be
%used for gesture and activity recognition, while providing a stable
%initialization for the shape estimation.  

%Meanwhile the shape
%estimation improves the pose accuracy by 
%provides dense spatial information which is useful
%for precise physical interaction, but also improves the 

% Paragraph three: what does this get us?

% Paragraph four: ???

%This provides accurate pose information which can be useful for
%gesture and activity recognition as well as dense spatial information
%which is necessary for more precise physical interaction.

%Template-based articulated tracking is an important problem in computer
%vision with applications in robotic perception, human robot interaction
%and performance capture. Mention multi-resolution.  Mention applications
%(esp. close range HRI, eg. shaving).

%Combining articulation and deformation gives you the low-d skeleton
%information which is good for things like activity recognition while
%also giving you high resolution surface information for more precise
%interaction.

%%%%%%%%%%%%%%%%%%%%%%%%%%%%%%%%%%%%%%%%%%%%%%%%%%%%%%%%%%%%%%%%%%
%%%%%%%%%%%%%%%%%%%%%%%%%%%%%%%%%%%%%%%%%%%%%%%%%%%%%%%%%%%%%%%%%%
\section{Related Work}
%%%%%%%%%%%%%%%%%%%%%%%%%%%%%%%%%%%%%%%%%%%%%%%%%%%%%%%%%%%%%%%%%%
%%%%%%%%%%%%%%%%%%%%%%%%%%%%%%%%%%%%%%%%%%%%%%%%%%%%%%%%%%%%%%%%%%
\label{sec:related_work}

%%%%%%%%%%%%%%%%%%%%%%%%%%%%%%%%%%%%%%%%%%%%%%%%%%%%%%%%%%%%%%%%%%
\subsection{Articulated Tracking and Pose Estimation}
%%%%%%%%%%%%%%%%%%%%%%%%%%%%%%%%%%%%%%%%%%%%%%%%%%%%%%%%%%%%%%%%%%

Articulated 3D tracking and markerless
motion capture
has been of interest to the computer
vision and
robotics communities for several years.  The objective of this
problem is to estimate the dynamic pose of a complex physical
object that can be parameterized using some low-dimensional
articulation structure, such as human bodies or hands.
Methods for solving this problem can be categorized
into discriminative methods which map directly from observations
to pose and generative methods which fit a model
to the observations, usually based on some previous estimate
or initialization.

%For example
%the Kinect skeleton tracker \cite{shotton2013real} was trained to
%directly detect model parts from a depth image using discriminative
%methods and combine those detections into a full body pose.
%On the other hand the DART tracker
%by Schmidt \cite{schmidt2014dart} and the
%work of Ye \cite{ye2014real} use gradient-based techniques
%to optimize an existing estimate of the model pose to fit new raw data.
%Both approaches have been heavily influenced by available 3D sensing
%hardware.

Discriminative methods
(\cite{haque2016towards, mehta2017vnect, shotton2013real})
and hybrid generative methods which incorporate
some discriminative component
(\cite{ganapathi2010real, helten2013personalization,
plagemann2010real, tompson2014real}) 
have the advantage that they are typically capable of single-frame pose 
estimation so that they can be initialized
automatically and can recover lost tracks easily in a video setting.
Recent methods also exist for tracking 3D volumes
without skeleton models using discriminative
approaches
\cite{huang2015toward, huang2016volumetric}, but
these also require significant training resources
and do not capture
high resolution surface detail.
In contrast generative methods for articulated model tracking
(\cite{chen1992object,  ganapathi2012real, GarciaCifuentes.RAL,
grest2005nonlinear, schmidt2014dart, ye2014real}) 
have the advantage that they can be readily applied to track new instances
or even entirely new classes of models so long as a template is available. 
For example the discriminative human-body tracker of
Shotten et al. \cite{shotton2013real} 
required collecting a massive amount of training data, a process which would have
to be repeated to track, for example, a dog. In contrast, Ye and Yang
showed that their generative technique for tracking human bodies translates
directly to tracking a dog by simply creating an appropriate
skeletal model \cite{ye2014real}.

Detecting and tracking human
pose from 2D images also has a long and important
history
(\cite{Cao2016RealtimeM2, carreira2016human, ferrari2008progressive, 
Iqbal2016PoseTrackJM}) and has gained significant attention
recently with the Microsoft COCO keypoint challenge
\cite{lin2014microsoft}.  While the goal of capturing human pose
is similar, this work is somewhat tangential to
our method as we aim to produce metrically accurate
3D estimation of pose and shape.

This work presents a new generative model-base tracking
technique which estimates fine-grained deformation of
the model mesh in addition to the skeletal pose.
Many articulated tracking methods, both generative and discriminative, 
have demonstrated robust performance when tracking subjects
such as human hands
and bodies that also exhibit non-rigid deformations.
However, the deformation is typically not modeled, which means
that information about surface shape is
not recovered. Some work
(\cite{de2008performance, gall2009motion}) attempt to model
this surface shape, but are not capable of online tracking and
real time performance.
Sometimes, as in the work of
Helten et al. \cite{helten2013personalization},
the template is initially adapted to a subject based on a set of images
of the subject in a canonical pose. Ye and Yang \cite{ye2014real} take
this a step further
by tracking the displacement of individual vertices in their mesh model
along the direction of the surface normal.  However, our hard
association of observed points to model vertices allows us reason
the entire mesh and point cloud in real time and therefore track much
finer details than is possible with the subsampled soft probabilistic
associations of Ye and Yang.
This in turn allows us to estimate much more fine-grained
deformations when used with a high-resolution mesh,
which is evident in the supplementary video.
%We also do not
%constrain vertex deformations in the direction of the surface normal,
%which is helpful for modeling deformations such as folds in clothing.

%Our method is
%similar to these approaches, but is distinct
%in two key ways.  First we forgo the more computationally expensive
%Gaussian mixture model of Ye and Yang in order to compute a
%higher resolution mesh.

%First our high resolution surface mesh is
%designed to capture fine details, while the adaptability of
%others is limited to overall body proportions.  Secondly our
%shape estimation is dynamic 

%our model shape dynamically adapts to fit
%fine detail and temporal deformation changes.

%Some systems have been built to track only one
%specific object category, while others are able to adapt to different
%categories using some specification of the object's structure.

%%%%%%%%%%%%%%%%%%%%%%%%%%%%%%%%%%%%%%%%%%%%%%%%%%%%%%%%%%%%%%%%%%
\subsection{Dynamic Surface Reconstruction}
%%%%%%%%%%%%%%%%%%%%%%%%%%%%%%%%%%%%%%%%%%%%%%%%%%%%%%%%%%%%%%%%%%
Mesh reconstruction of dynamic scenes has also been
of interest for some time.  For example Li et al.
\cite{li2012temporally} present this as a temporally
coherent shape completion on meshes with only partial
observability.  
Recent papers have shown that it is possible to perform 3D mesh
reconstruction of dynamic scenes in real time.
Initially these methods required a
model scanning phase before tracking
\cite{zollhofer2014real} but later work
dropped this requirement
\cite{dou2016fusion4d, innmann2016volume, newcombe2015dynamicfusion}.
These recent methods work by using
reconstruction techniques such
as volumetric SDF fusion \cite{curless1996volumetric}
while simultaneously estimating deformation
parameters that warp
the reconstructed mesh into its current shape.
Reconstruction techniques have the advantage that
they can produce
accurate shape information with no template and
no training data
required beforehand.
%The use of reconstruction techniques ensures that the
%resulting mesh will be well suited to a variety of different objects
%without needing to construct specific models beforehand.
However, starting each reconstruction from scratch results in a lack of
correspondence between multiple reconstructions of the same subject.
In contrast, we can identify correspondences through our template model
within and across video sequences.
Furthermore, non-rigid reconstruction techniques must rely on deformation
models that are general enough to capture any possible deformations, from
fully non-rigid objects such as towels to skeleton-based models such as
human bodies.
By making use of model-specific prior knowledge, our technique is able
to track the majority of the motion in a much lower dimensional pose space,
making the optimization more efficient as well as
provide the resulting pose as useful additional data.

%%%%%%%%%%%%%%%%%%%%%%%%%%%%%%%%%%%%%%%%%%%%%%%%%%%%%%%%%%%%%%%%%%
%%%%%%%%%%%%%%%%%%%%%%%%%%%%%%%%%%%%%%%%%%%%%%%%%%%%%%%%%%%%%%%%%%
\section{Method: Template Model}
%%%%%%%%%%%%%%%%%%%%%%%%%%%%%%%%%%%%%%%%%%%%%%%%%%%%%%%%%%%%%%%%%%
%%%%%%%%%%%%%%%%%%%%%%%%%%%%%%%%%%%%%%%%%%%%%%%%%%%%%%%%%%%%%%%%%%
\label{sec:method_template_model}

%%%%%%%%%%%%%%%%%%%%%%%%%%%%%%%%%%%%%%%%%%%%%%%%%%%%%%%%%%%%%%%%%%
%\subsection{Overview}
%%%%%%%%%%%%%%%%%%%%%%%%%%%%%%%%%%%%%%%%%%%%%%%%%%%%%%%%%%%%%%%%%%

%Articulated tracking is the problem of fitting a jointed kinematics model
%to a stream of live video observations.  Our method also dynamically updates
%a mesh surface representation of the tracked subject.  In this section
%we detail our approach to this problem.
Similar to Ye and Yang \cite{ye2014real} and
Schmidt et al. \cite{schmidt2014dart}
our technique uses an
iterative gradient-based approach to fit a kinematics model
to the observed data. We assume that the tracking sequence
starts with an initial estimate of the skeletal pose.
From that initialization, we iteratively optimize
the pose to fit each incoming frame and then use a second
optimization to update the vertex positions of a triangle mesh
representing the object's surface.  Section
\ref{subsec:kinematic_structure} explains the kinematic
structure of our skeleton, while
\ref{subsec:surface_representation} and
\ref{subsec:dynamic_shape_parameters} detail the
model's shape representation.
After this, Section \ref{sec:method_optimization}
explains the optimization processes used to fit this model
to live data.

%%%%%%%%%%%%%%%%%%%%%%%%%%%%%%%%%%%%%%%%%%%%%%%%%%%%%%%%%%%%%%%%%%%%
\subsection{Dual Quaternion Kinematic Structure}
%%%%%%%%%%%%%%%%%%%%%%%%%%%%%%%%%%%%%%%%%%%%%%%%%%%%%%%%%%%%%%%%%%%%
\label{subsec:kinematic_structure}

\begin{figure}[t]
\centering
\includegraphics[width=0.45\textwidth]{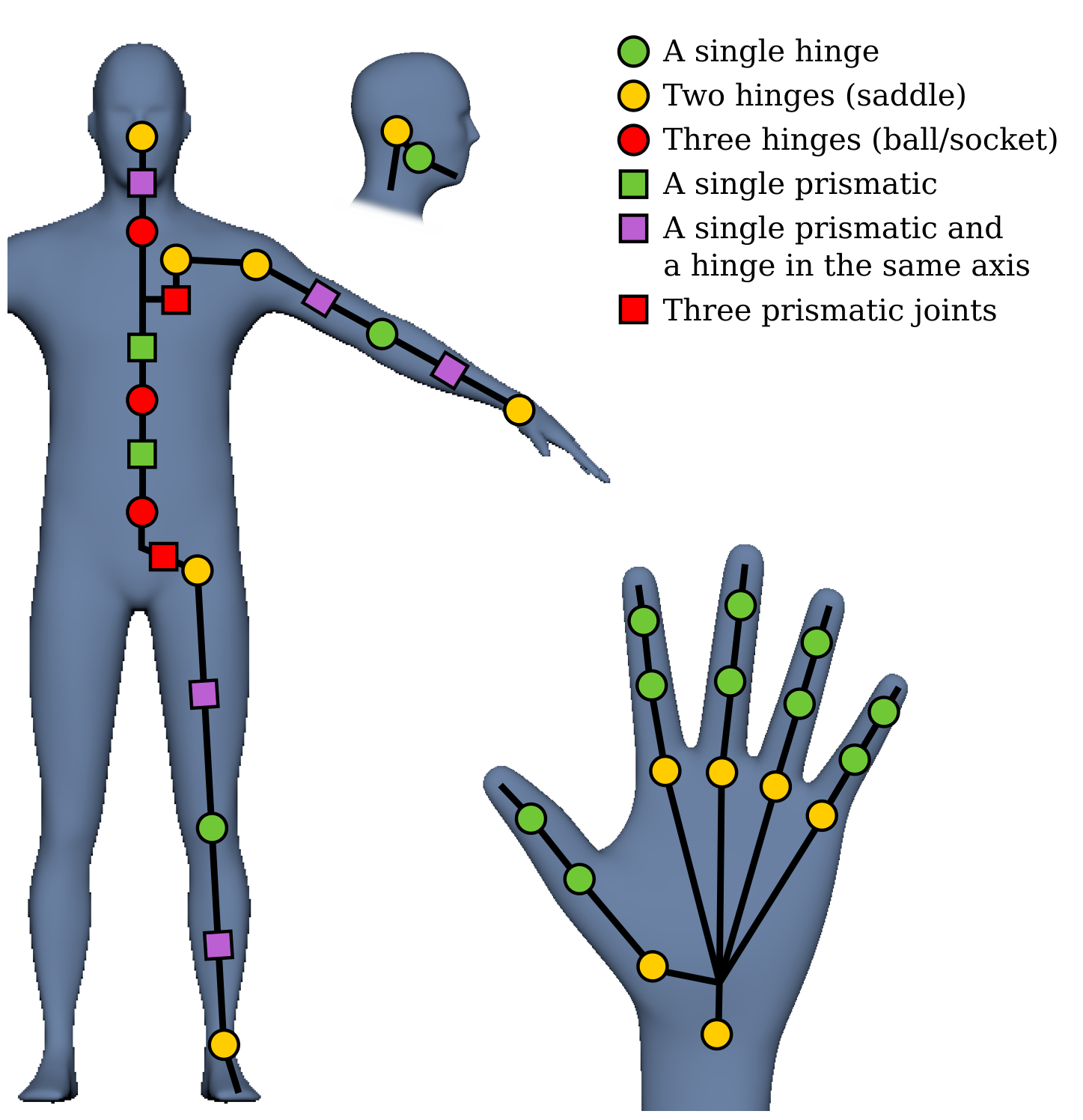}
\caption{The joint layout of our model.}
\label{fig:joint_layout}
\end{figure}

Our model consists of a skeleton with an attached
mesh.
%The skeleton is similar to the kinematics of a robotic
%manipulator and
The skeleton is made of link frames connected by hinge (rotation)
and prismatic (translation) joints.
While our human model is primarily made up
of hinge joints, we use
some prismatics to allow subtle stretching in order
to fit subjects with varying proportions and correct
for subtle modeling or joint placement errors.
%We find in practice that this greatly aids tracking
%performance.
Joints such as the
shoulder with more than one rotational degree
of freedom are represented as multiple hinge joints
in succession.
%The skeleton forms a tree in which each link frame
%is connected to
%at least one other link by a hinge or prismatic joint.
The kinematic hierarchy of our human model is shown
in Figure~\ref{fig:joint_layout}.

We use the dual quaternion parameterization of SE(3),
originally proposed by Clifford
\cite{clifford1882mathematical}, to represent the position
and orientation of each link in the hierarchy.
While this representation may be less familiar,
Kavan et al. \cite{kavan2008geometric} has shown that it
can be used to provide superior performance for smooth
mesh attachment.  This is discussed later in Section
\ref{subsec:surface_representation}.
Dual quaternions consist of two quaternions of the form
$H = q_r + q_d \epsilon$.  Here $\epsilon$ refers to
Clifford's dual unit which satisfies $\epsilon^2 = 0$.
The first quaternion $q_r$ is referred to as
the \textit{real}
part and represents the rotational
component of the transformation while $q_d$ is
the \textit{dual} part and represents
translation.
For the sake of space, we omit a thorough coverage of the
mathematical details and instead refer readers to
\cite{daniilidis1999hand}.

%%%%%%%%%%%%%%%%%%%%%%%%%%%%%%%%%%%%%%%%%%%%%%%%

Hinge joints are parameterized by a unit axis
$z \in \mathbb{S}^2$, and an angle value
$\theta \in \mathbb{R}$.  Prismatic joints are similarly
parameterized by a unit axis $z \in \mathbb{S}^2$ and
a translation value $\theta$.
In our model, the axes $z$ are fixed
while $\theta$ changes over time to represent the model pose.
%Figure \ref{fig:joint_parameters} shows three links
%connected by a hinge and a prismatic joint.
Using dual quaternions, a hinge transformation
$H_{h}(\theta; z)$
representing a rotation by $\theta$ about a fixed
axis $z$ has the form:
\begin{equation}
\label{eq:H_hinge}
H_{h}(\theta; z) = \cos(\theta/2) +
(z_x i + z_y j + z_z k) \sin(\theta/2) + 0 \epsilon
\end{equation}
%\[
%H_{r(\theta,z)} = \cos \left( \theta/2 \right) +
%\langle z, [i,j,k] \rangle
%\sin \left( \theta/2 \right) + 0 \epsilon
%\]
%\[
%H_{r(\theta,z)} = \cos \left( \theta/2 \right) +
%z^T \begin{bmatrix}i\\j\\k\end{bmatrix} \sin \left( \theta/2 \right) %+ 0 \epsilon
%\]

%where $i,j$ and $k$ are the quaternion units and
%where the angle brackets
%$\langle,\rangle$ signify the dot product. 
A prismatic transform representing a translation
by $\theta$ along an axis $z$ can be constructed as:
\begin{equation}
\label{eq:H_prismatic}
H_{p}(\theta;z) =
%1 + \langle z, [i,j,k] \rangle \frac{\theta}{2} \epsilon
1 + (z_x i + z_y j + z_z k) \frac{\theta}{2} \epsilon
\end{equation}
%In order to build a hinge joint with a pivot we need to first
%subtract the pivot, then rotate, then add the pivot again:
%\begin{equation}
%\label{eq:H_hinge}
%H_{h}(\theta;z,p) = H_t(p) H_r(\theta;z) H_t(-p)
%\end{equation}
%where $H_t(x)$ represents a translation by vector $x$.

%With this machinery in place we can construct a set of equations for
%the link transformations relative to the camera frame.
%The root link, which we assign index 1, is able to
%freely translate and rotate relative to the camera frame $c$,
%so we store this offset explicitly as a dual quaternion
%$H_{c,1}$.
We also store a fixed offset
$H_{\mathcal{P}(j),j} \in SE(3)$ between
each joint and the
link's origin.  This allows us to specify a pivot point
and realign the axes if convenient.
The link frames are arranged in a hierarchy so we
can compute %subsequent
the offset between the world frame and any link frame
using the recursive definition:
\[
H_{0,j} = H_{0,\mathcal{P}(j)} H_{\mathcal{P}(j),j}
H_j
\]
where $H_{0,0}$ is the identity matrix, $H_j$ is the
joint transform connecting link $j$ to its direct
parent $\mathcal{P}(j)$, and is either a hinge or
prismatic joint.

%The root link, indexed as 1, is able to freely translate and rotate
%relative to the world/camera frame, so we store this
%offset explicitly as $H_{c,1}$ as another dual quaternion.
%The link frames are arranged in a hierarchy so we can use the
%equations for the local offsets between frames to construct the
%transform from the camera to any link $j$ by the recursive definition
%\[
%H_{c,j} = H_{c,1} H_{1,parent(j)} H_{parent(j),j}
%\]

%For convenience of notation we refer to
%the transform of link $j$ relative to the world as $L_j = H_{c,j}$
%and $|L|$ as the number of links.

%The links frames in our model are arranged in a hierarchy,
%so we can use the equations for the local offsets between frames
%to construct the transform between any frame $j$ and the root frame
%$H_{1,j}$ by the recursive definition
%\[
%H_{1,j} = H_{1,parent(j)} H_{parent(j),j}
%\]
%where $H_{0, parent(j)}$ is the transform between the direct parent of
%frame $j$ and the root frame and $H_{parent(j),j}$ is a hinge or
%prismatic transform.

%%%%%%%%%%%%%%%%%%%%%%%%%%%%%%%%%%%%%%%%%%%%%%%%%%%%%%%%%%%%%%%%%%%%
\subsection{Surface Representation and Smooth Skinning}
%%%%%%%%%%%%%%%%%%%%%%%%%%%%%%%%%%%%%%%%%%%%%%%%%%%%%%%%%%%%%%%%%%%%
\label{subsec:surface_representation}

\begin{figure}[t]
\centering
\includegraphics[width=0.48\textwidth]{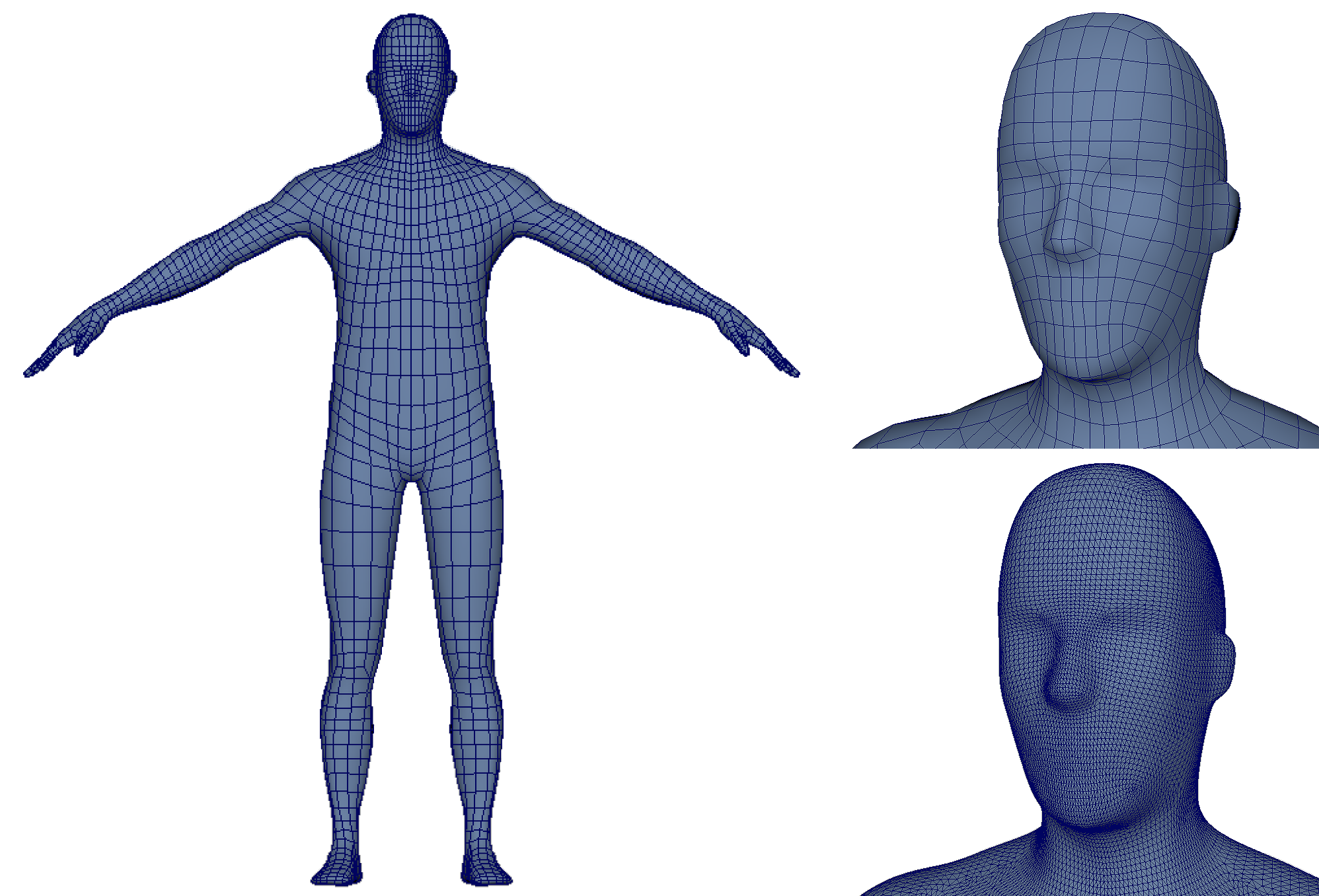}
\caption{Our human body mesh.  The left and top right show the low
resolution version while the bottom right shows the smoothed high
resolution version.}
\label{fig:mesh}
\end{figure}

The model's surface is represented as a high resolution triangle mesh.
Unlike the kinematic skeleton, we do not assume any initial estimate of
this mesh and initialize it to a smooth default shape shown in figure
\ref{fig:mesh}.
This mesh consists of set of 3D vertex positions
$V = \left\{v_1 \hdots v_{|V|} \right\},\ v_i \in \mathbb{R}^3$
as well as a triangle list $F$.
Each triangle is represented as a set of three integers referencing
vertex indices
$F = \left\{f_1 \hdots f_{|F|} \right\},
\ f_k \in \mathbb{Z}^3$.
This way the model can transform the mesh by updating
the vertex positions while leaving the face list fixed.
The mesh is attached to the skeleton using dual quaternion
skinning \cite{kavan2008geometric}.
This provides a way to smoothly blend the influence of links
between different regions of the mesh.
%This technique is also
%used by Dynamic Fusion \cite{newcombe2015dynamicfusion} to attach
%the reconstructed mesh to a set of warp node transforms.
%Figure \ref{fig:skin_blending} shows a diagram of the improvements
%of smooth skinning over split segments and rigid assignment of
%vertices to links.
Dual quaternion skinning requires a bind pose $H_{0,j}^0$
for the skeleton link frames, as well
as a weight matrix $\Omega$.
The bind pose represents the pose for which the
kinematic skeleton matches the default pose of the mesh.
We build our skeleton so that the pose in which $\theta=0$
for all joints is the bind pose.
The weight matrix describes the influence of each frame on each vertex.
Each column $\omega_i$ corresponding to vertex $i$ is constrained
such that
\[
\omega_i \in \left[0,1\right]^{|L|},\ \sum_{j=1}^{|L|} \omega_{ij} = 1
\]
where $|L|$ is the number of skeleton links.
Most vertices are weighted to only one or two joints, so
we limit the number of non-zero entries in each $\omega_i$
to be four and use a sparse representation to store this data
in order to limit memory overhead.

%%%%%%%%%%%%%%%%%%%%%%%%%%%%%%%%%%%%%%%%%%%%%%%%
%%%%%%%%%%%%%%%%%%%%%%%%%%%%%%%%%%%%%%%%%%%%%%%%

Given this information the skinning function transforms a vertex
by computing the offset between the bind pose and the current pose
of each frame and then constructing a linear blend $H_{i \Sigma}$
of these offsets for each vertex $v_i$ based on the weights.
\begin{equation}
\label{eq:H_Sigma}
H_{i \Sigma} =
\left(\sum_{j=1}^{|L|} H_{0,j} \left(H^0_{0,j}\right)^{-1} 
\omega_{ij} \right)
\end{equation}

The skinned vertex position $v_i$ can be computed by multiplying
this transformation by the vertex position in the default model
$v^0_i$.
\begin{equation}
\label{eq:skin}
v_i = H_{i \Sigma} v^0_i
\end{equation}

%Note that $H_{i\Sigma}$ in equation \ref{eq:H_Sigma} is
%a linear combination of dual quaternions and is therefore not
%guaranteed to have a unit norm.  It is possible to perform a
%normalization step to correct this, but because we are only
%transforming 3D points and not orientations or transformations,
%this normalization does not change the result and can be omitted.

%but this would only be
%necessary if $H_{i\Sigma}$ were used to multiply another
%dual quaternion.  The equations for transforming a 3D point
%are unaffected by omitting this step.

%Technically $T_{i\Sigma}$ in equation \ref{eq:TSigma} above
%should include a dual quaternion normalization in order for the
%resulting dual quaternion to represent an element of $SE(3)$.
%Fortunately the normalization is only important when the result will be
%used to multiply another dual quaternion and the equations for
%transforming 3D points are unaffected by omitting it.

%%%%%%%%%%%%%%%%%%%%%%%%%%%%%%%%%%%%%%%%%%%%%%%%%%%%%%%%
\subsection{Dynamic Shape Parameters}
%%%%%%%%%%%%%%%%%%%%%%%%%%%%%%%%%%%%%%%%%%%%%%%%%%%%%%%%
\label{subsec:dynamic_shape_parameters}

Dynamic shape deformation is represented by a set of warp
offsets $\Phi$
containing a three vector
$\phi_i \in \mathbb{R}^3$ for each vertex describing a translation
away from it default position.
%This makes the deformed position of $v_i$:
We can augment Equation \ref{eq:skin} above to
compute the warped position $v_i$:
\begin{equation}
\label{eq:deformation}
v_i = H_{i \Sigma}(v^0_i + \phi_i)
\end{equation}
%In order to account for dynamic shape deformation,
%we store a vector $\phi_i \in \mathbb{R}^3$
%for each vertex describing it's offset from the default position it would
%have due to skeletal transformation.

%The mesh vertex positions can be computed as a function of the pose
%$\theta$, the weights $\omega$ and the vertex offsets $\phi$ as
%\begin{equation}
%\label{eq:skin}
%v_i = \left(\sum_{j=1}^{n_f} T_{0,j} T^0_{j,0} \omega_{ij} \right)
%(v^0_i + \phi_i)
%\end{equation}

%In addition to the influence of the joints we store a translation offset
%$u_i$ for each vertex independently in order to capture dynamic shape
%deformation.
%The mesh vertex positions can be computed as a function of the pose
%$\theta$, the weights $\omega$ and the vertex offsets $u$ as
%\[
%v_i = \left(\sum_{j=0}^{n_f} q_{0,j} (q^0_{j,0}) \omega_{ij} \right)
%(v^0_i + u_i)
%\]
%where $q^0_{j,0}$ is the transform between the origin and the
%default pose of frame $j$, and $v^0_i$ is the default position
%of the mesh vertex.

In order to capture high frequency shape details, the mesh
necessarily contains a very large number of vertices and triangles.
Unfortunately large meshes are unwieldy and it can be difficult
to generate the skin weights for them
effectively.  To avoid this we worked with a low
resolution polygon mesh containing 3,460 vertices and 3,476 faces.
We then generated a high resolution version automatically using
two iterations of Catmul-Clark subdivision
\cite{catmull1978recursively}.  After triangulating the
resulting quadrilaterals, this resulted in a mesh with
55,474 vertices and 110,994 triangles.  Figure \ref{fig:mesh}
shows the low resolution and high resolution meshes.
We also
generated high resolution skin weights from the low resolution
mesh by interpolating them using the same scheme that
Catmul-Clark subdivision uses to interpolate
vertex positions.

%\textbf{Constructing our model using multiple resolutions
%also allowed us to do coarse-to-fine data association which
%is described in section \ref{subsec:data_association}.}
%As a final note, we store connections between vertices
%corresponding to the original low resolution mesh
%and the interpolated high resolution vertices that they produce.
%These connections are used later for interpolating shape error
%in section \ref{subsec:shape_optimization}.

The low resolution mesh originated from a human model
from the website CG Trader \cite{cgtrader}
by the NoneCG group~\cite{nonecg} and was used with
permission.  We heavily modified this mesh and
constructed the skeleton
hierarchy and skin weights by hand
using Autodesk's Maya software~\cite{maya2016}.

%%%%%%%%%%%%%%%%%%%%%%%%%%%%%%%%%%%%%%%%%%%%%%%%%%%%%%%%
%\subsection{Pose and Shape Estimation}
%%%%%%%%%%%%%%%%%%%%%%%%%%%%%%%%%%%%%%%%%%%%%%%%%%%%%%%%
%\label{subsec:optimization}

%%%%%%%%%%%%%%%%%%%%%%%%%%%%%%%%%%%%%%%%%%%%%%%%%%%%%%%%
%%%%%%%%%%%%%%%%%%%%%%%%%%%%%%%%%%%%%%%%%%%%%%%%%%%%%%%%
\section{Method: Optimization}
%%%%%%%%%%%%%%%%%%%%%%%%%%%%%%%%%%%%%%%%%%%%%%%%%%%%%%%%
%%%%%%%%%%%%%%%%%%%%%%%%%%%%%%%%%%%%%%%%%%%%%%%%%%%%%%%%
\label{sec:method_optimization}

Our model fitting approach alternates between optimizing
the skeleton pose and the dynamic warp parameters.  Section
\ref{subsec:data_association} explains the residual term
we use for fitting while Section
\ref{subsec:kinematic_optimization} and
\ref{subsec:shape_optimization} discuss kinematic
optimization and shape optimization respectively.

%%%%%%%%%%%%%%%%%%%%%%%%%%%%%%%%%%%%%%%%%%%%%%%%%%%%%%%%
\subsection{Data Association and Residual Term}
%%%%%%%%%%%%%%%%%%%%%%%%%%%%%%%%%%%%%%%%%%%%%%%%%%%%%%%%
\label{subsec:data_association}

Given the model described in Section
\ref{sec:method_template_model}
the task of estimating pose and shape requires estimating
the joint angles $\Theta$ and the vertex offsets $\Phi$.
This is achieved by first generating a residual term describing
the offset between the model and the observations, computing
the derivative of that residual with respect to the parameters
$\Theta$ and $\Phi$ and then solving a linear system
to compute an update that reduces the
residual.

Our observations take the form of a point cloud
$P = \left\{p_1 \hdots p_{|P|} \right\}
\ p_k \in \mathbb{R}^3$.  Because we use a depth camera
to capture these point clouds, each point corresponds to
a pixel in a two-dimensional grid.
%Constructing an
%association between these observations and the model is
%of critical
%importance to articulated tracking algorithms using a
%template model.
Data association techniques are an important differentiating
factor in template based trackers.
Ye and Yang \cite{ye2014real} use a
Gaussian Mixture Model with centroids at the mesh vertex
positions to explain the data.  While they report that
this performs well, this computation is expensive and requires
that they subsample their mesh and point cloud when computing
the association.  Given our high resolution model and our goal
of accurate detailed shape estimation, this method would
not be feasible in our system.
Schmidt et al. \cite{schmidt2014dart} use 
signed distance functions generated from
rigid link geometry to compute association.  Unfortunately
this is also infeasible in our case because we use a
single non-rigid mesh to represent the entire subject.
Computing a new signed distance function for this mesh
for every optimization update would be too slow for
our purposes.

Instead of the methods above, we use projective data
association that utilizes the grid structure of the
point cloud to perform nearest neighbor search.
%This involves using the known camera
%parameters to project the three-dimensional vertex
%positions onto the two dimensional pixel grid and
%assigning observation points to vertices by searching
%in a small window.
%We start by constructing a bin for each pixel in the depth
%image.  We then project each three dimensional vertex
%onto this grid using the intrinsic camera parameters and
%assign them to the pixel bins.  We then iterate over every
%point in the observed point cloud and search in a small
%window 
We start by projecting each three dimensional vertex onto
the image plane and placing them into buckets corresponding to
pixels.  We then iterate through all points in the observed
point cloud and exhaustively search the buckets in a window
around the corresponding pixel for the closest vertex, ignoring
anything that is farther away than a cutoff threshold.
This guarantees that the closest vertex will be found as long as
the window size is chosen correctly relative to the subject's
minimum distance to camera.

At this point it is possible that many point cloud observations
have been assigned to the same vertex, so we average the three
dimensional offset between the vertex and each point that has been
assigned to it.  We then compute a point plane residual using this
offset and the model's vertex normal \cite{chen1992object}.
%\textbf{Add residual interpolation.}
%\textbf{Figure out how to make this all about residuals.}
%The residual is computed by first constructing an
%assignment between
%vertices and observations.
%Because we only have observations for the camera-facing side of
%the model we do not perform data assignment for back-facing
%vertices and set their contribution to the error to be zero.
%Once we have the assignment we compute a
%point-plane residual term \cite{chen1992object} for each
%vertex based on these sources and interpolate them.
%Because there may be multiple observation points for
%which a given vertex is closest, we average these
%points together when computing the residual for each vertex.
If we let $N = \left\{ n_1 \hdots n_{|V|} \right\}$ be the vertex normals,
%$\widehat{P} = \left\{
%\widehat{p}_1 \hdots \widehat{p}_{|V|} \right\}$
%be the closest observed point to each vertex
and $\widetilde{P} = \left\{ \widetilde{p}_1 \cdots\widetilde{p}_{|V|}
\right\}$ be the average of the observation points
for which each vertex in $V$ is the closest, the residual
term for each vertex is

%\begin{equation}
%\label{eq:residual}
%r_{data} = \sum_{i=1}^{|V|} n_i^T
%\left(\lambda_{m} (\widehat{p}_i - v_i) +
%(1 - \lambda_{m}) (\widetilde{p}_i - v_i)
%\right)
%\end{equation}

%\begin{equation}
%\label{eq:residual}
%r_{data} = \sum_{i=1}^{|V|} n_i^T
%(\widetilde{p}_i - v_i)
%\end{equation}

\begin{equation}
\label{eq:residual}
r_i = n_i^T (\widetilde{p}_i - v_i)
\end{equation}

The goal of our optimization procedure is to reduce the sum of the
squares of these residual terms.  In many ways this is a simpler and
more direct association technique than that used by Ye and Yang and
Schmidt et al.  Indeed, as we show in Section \ref{sec:experiments},
we find that it does not perform as well on
noisy low resolution sequences such as the EVAL dataset.
However on high resolution sequences from a modern time of flight
sensor this is more than adequate and has the advantage that it
is extremely fast to compute and does not require any complex
data structures.  This allows us to operate on a very high
resolution mesh in real time, which was not possible with previous
approaches.

\subsection{Kinematic Optimization}
%%%%%%%%%%%%%%%%%%%%%%%%%%%%%%%%%%%%%%%%%%%%%%%%%%%%%%%%%%%%%%%%%%
\label{subsec:kinematic_optimization}

% Need to try it the other way again and see if anything is better
%The skeleton is optimized without taking the dynamic deformation into
%account.  This is equivalent to assuming the $\phi_i$
%values are zero.
%This is done to prevent the mesh from drifting off the skeleton
%over time.
The residual term in Equation \ref{eq:residual} is a function
of the vertex positions $V$ and
the position of each vertex is a function of the skeleton pose
$\Theta$, the bind
pose $H^0$, the default mesh vertices $V^0$, the vertex offsets
$\Phi$ and the skin weights $\Omega$.
This means we can optimize the skeleton pose by computing the
gradient of the residual with respect to the joint values $\Theta$
and using damped least squares \cite{marquardt1963algorithm}
to take an optimization step that reduces the sum of the squares
of these residuals.
%If $f_{skin}$ is
%the skinning function from Equation \ref{eq:deformation} and
%$f_{data}$ is the residual function from Equation \ref{eq:residual}
%we can write:
%\[
%e_{pose} = f_{data}(f_{skin}(V^{0},H^{0},\Omega,\Theta, \Phi))
%+ \lambda_{s} ||S \Theta ||_2^2
%\]
%where the last term is a default pose prior.
This requires a Jacobian expressing the derivative of the
residual $r_i$
%$e_{pose}$
with respect to the joint angles $\Theta$.  For a single vertex
we can write this as:
\[
%\frac{\partial e_{pose}}{\partial \Theta} =
%\frac{\partial e_{pose}}{\partial V} \frac{\partial V}{\partial \Theta}
%+ \lambda_{s} \frac{\partial ||S \Theta ||^2_2}{\partial \Theta}
\frac{\partial r_i}{\partial \Theta} =
\frac{\partial r_i}{\partial v_i} \frac{\partial v_i}{\partial \Theta}
\]
Because we use point-plane error, the derivative of the residual with
respect to each component of the vertex position is simply the vertex
normal.  The derivative of the vertex position with respect to
the skeleton pose $\Theta$ is more complex, but can still be computed
analytically.  Recall that the skinning operation transforms a model
vertex from its offset mesh position $(v^0_i + \phi_i)$ to it's posed
position $v_i$ by multiplying it by the blended dual quaternion 
$H_{i\Sigma}$ from Equation \ref{eq:H_Sigma}.
We can use $H_{i\Sigma}$ and $H_{j\Delta} = H_{0,j}
\left(H^0_{0,j} \right)^{-1}$
as intermediate variables
and write:
\[
%\frac{\partial V}{\partial \Theta} =
%\frac{\partial V}{\partial H_{\Sigma}}
%\frac{\partial H_{\Sigma}}{\partial H_{\Delta}}
%\frac{\partial H_{\Delta}}{\partial \Theta}
\frac{\partial v_i}{\partial \Theta} =
\frac{\partial v_i}{\partial H_{i\Sigma}}
\frac{\partial H_{i\Sigma}}{\partial H_{\Delta}}
\frac{\partial H_{\Delta}}{\partial \Theta}
\]

The first term describes how the three dimensional vertex position
$v_i$ changes with respect to the eight-dimensional
blended dual quaternions $H_{i\Sigma}$ as a
%$3 |V|$ by $8 |V|$
3 by 8 matrix.

%The second term describes how the blended transforms change with
%respect to the link offsets $H_{\Delta}$
%as an %$8|V|$
%8 by $8 |L|$ matrix.  The last term describes
%how the link offsets change with respect to the joint pose
%parameters as a $8 |L|$ by $|\Theta|$ matrix.  We can now
%examine each of these terms individually.

%Each vertex is only affected by a single $H_{\Sigma}$, so the
%first matrix $\partial V / \partial H_{\Sigma}$ will be block
%diagonal and can be stored
%compactly using $24|V|$ floats.
%Differentiating each component of $v_i$ with respect to each
%component of $H_{i\Sigma}$ results in a 3x8 partial derivative
%matrix.  We omit the full matrix here for space considerations, but
%it can be computed by differentiating the equation for transforming a
%point by a dual quaternion found in \cite{daniilidis1999hand}.

$H_{i\Sigma}$ is a linear combination of the link offsets
$H_{\Delta}$ for vertex $i$ using the $\Omega$ weight matrix.
This means that the weight matrix can be used directly to construct
$\partial H_{i\Sigma}/\partial H_{\Delta}$ as an $8$ by $8 |L|$ matrix.
Each $8\times 8$ block is simply the identity matrix scaled by
$\omega_{ij}$.

%To compute the second term $\partial T_{\Sigma} / \partial \theta$
%recall that $T_{\Sigma}$ is a weighted sum of dual quaternions.
Because of the hierarchical nature of the skeleton, a single 
link can be influenced by several $\theta$ values.
For example the spine, shoulder and elbow joints will all
influence the transform of the hand link.
%To see this
%note that the $T_{0,j}$ term in equation \ref{eq:H_Sigma} is
%recursively defined as $T_{0,parent(j)} T_{parent(j),j}$.
Even though we restrict $\omega$ so that only four
frames can influence a single vertex, those four frames may be
influenced by many joints in the kinematic hierarchy, which
means that the $\partial H_{\Delta} / \partial \Theta$ matrix
is relatively dense. 
If $\theta_k$ is an ancestor of link $j$ in the hierarchy, we can
compute a block of this matrix
corresponding to $H_{j\Delta}$ and $\theta_k$ as
\[
H_{0,\mathcal{P}(k)}
H_{\mathcal{P}(k),k}
\frac{\partial H_k}{\partial \theta_k} H_{k,j}H_{j\Delta} %~,
\]
Otherwise if $\theta_k$ is not an ancestor link $j$ then
this block will be zero.
%%%%%
%where $k$ is the index of the link defined by a prismatic
%or hinge joint with state dependent on $\theta_k$.
%%%%%
If $\theta_k$ corresponds to a
prismatic transform from Equation
\ref{eq:H_prismatic}, its derivative is
\[
%\frac{\partial H_{\mathcal{P}(i),i}(\theta_k;z)}{\partial \theta_k} =
\frac{\partial H_k}{\partial \theta_k} = 
\frac{ (z_x i + z_y j + z_z k)}{2}\epsilon
\]
If $\theta_k$ corresponds to a hinge transform from
Equation \ref{eq:H_hinge}, the derivative is
\[
%\begin{equation}
%\begin{split}
%\frac{\partial H_{\mathcal{P}(i),i}(\theta_k;z,p)}
%{\partial \theta_k} = \\
\frac{\partial H_k}{\partial \theta_k} =
%H_t(p)
\frac{ (z_xi + z_yj + z_zk) \cos(\theta_k/2) -
\sin(\theta_k/2)}{2}
%H_t(-p).
%\end{split}
%\end{equation}
\]

%We can compute a block of this matrix
%corresponding to $H_{j\Delta}$ and $\theta_k$ as
%\[
%H_{1j}
%\frac{\partial H_\theta}{\partial \theta} H_{1j}^{-1} H_{j\Delta}
%\]
%where $H_\theta$ corresponds to either a
%prismatic transform $H_{t(\theta,z)}$ from Equation
%\ref{eq:H_prismatic} or a hinge transform
%$H_{h(\theta,z,p)}$ from Equation \ref{eq:H_hinge}.
%For a prismatic joint, the
%$\partial H_{t(\theta,z)} / \partial \theta$ term is
%\[
%\frac{\partial H_{t(\theta,z)}}{\partial \theta} =
%\frac{\langle z, [i, j, k] \rangle \epsilon}{2}.
%\frac{ (z_x i + z_y j + z_z k)}{2}\epsilon
%\]
%For a hinge joint it is
%\[
%H_{parent}
%\frac{\partial H_{h(\theta,z,p)}}{\partial \theta} =
%\frac{\langle z, [i, j, k] \rangle \cos(\theta/2) -
%\sin(\theta/2)}
%{2}.
%\]
%\textbf{There is some small inaccuracy above.  For the hinge
%joint in this configuration, $\theta$ is always zero.  To fix
%this, either $H_{1j}$ or $H_{1j}^{-1}$ needs to be to the parent
%joint, AND the pivot needs to be incorporated.  Either explain that
%this is going on or fix it.}

The damped least squares method
involves solving 
\[
(J^TJ+\lambda_{k} \text{diag}(J^TJ))x = J^Tr
\]
for x.  The full $J^TJ$ and $J^Tr$ matrices can be computed as
\begin{equation}
\label{eq:JTJ}
J^TJ = \sum_{i=1}^{|V|}
\left(\frac{\partial r_i}{\partial \Theta} \right)^T
\frac{\partial r_i}{\partial \Theta}
\end{equation}
\begin{equation}
\label{eq:JTr}
J^Tr = \sum_{i=1}^{|V|}
\left(\frac{\partial r_i}{\partial \Theta} \right)^T
r_i
\end{equation}
For a high resolution mesh, this can be efficiently computed
on a GPU by computing each
$(\partial r_i/\partial \Theta)^T \partial r_i/\partial \Theta$
in parallel and using atomic operations to sum them.

As a final addition, we add a default pose prior that penalizes the
squared value of $\theta$ for all joints.  This encourages the
optimization to relax towards the default pose when there are few
observations for a particular link and avoid joint limits.
The values of $\Theta$ are multiplied by a diagonal
matrix $S$ that weights each joint by the number of vertices it
influences.  This prevents the penalty from overwhelming smaller
joints that might not get enough observations to overcome it
otherwise.  To do this we augment Equations \ref{eq:JTJ} and
\ref{eq:JTr}:
\[
J^TJ = \sum_{i=1}^{|V|}
\left(\frac{\partial r_i}{\partial \Theta} \right)^T
\frac{\partial r_i}{\partial \Theta}
+
(\lambda_s S)^2
\]
\[
J^Tr = \sum_{i=1}^{|V|}
\left(\frac{\partial r_i}{\partial \Theta} \right)^T
r_i
+
(\lambda_s S)^2 \Theta
\]

%Finally we must include the contribution of the default pose
%prior.  This penalizes the squared value of $\theta$
%for all joints.  The purpose of this penalty is to
%encourage the optimization to relax towards the default pose
%when there are few observations for a particular link and to avoid
%joint limits.  The values of $\Theta$ are multiplied by a diagonal
%weighting matrix $S$ before this penalty is applied.  This matrix
%weights each joint by how many vertices it influences.  This
%prevents the penalty from overwhelming smaller joints, for example
%in the fingers, that might not get enough observations to overcome
%it otherwise.  The derivative of this component is
%\[
%\lambda_{s} \frac{\partial||S\Theta||^2_2}{\partial \Theta} =
%2 \lambda_{s} S \Theta
%\]

%Once we have these matrices we combine them into a single
%Jacobian $J = \partial r / \partial \Theta$ representing the
%change in residual with respect to each
%joint pose $\theta$.  We then compute a pose offset using
%damped least-squares \cite{marquardt1963algorithm}.  This
%involves solving
%\[
%(J^TJ+\lambda_{rk} \text{diag}(J^TJ))x = J^Tr
%\]
%for $x$.
The $(J^TJ + \lambda_{k} \text{diag}(J^TJ))$ matrix
is positive semi-definite, meaning the system can be solved efficiently
using Cholesky decomposition \cite{gill1974newton} implemented in
CUDA.  Once $x$ has
been computed it is subtracted from the current pose $\Theta$ and
the process is repeated.  As the pose is updated, the smooth
skinning operation pulls the mesh into place and provides an
initialization point for the shape optimization.
In practice we have found that ten to
fifteen iterations of this optimization for each incoming frame
is sufficient to match the pose of the target and keep the system
running at real time frame rates.  The top right
frame of Figure \ref{fig:one} shows the result of fitting the
kinematic model with the default mesh onto a point cloud without
additional shape estimation.

%There is one additional consideration that must be taken into
%account.  The mesh vertices are able to freely move via
%the $\Phi$ parameters so steps must be taken to ensure that they
%don't collectively drift away from the underlying skeleton.
%There are two mechanisms that prevent this.  First, instead of
%using the true warped vertex positions $V$ in our computation of
%the skeleton dynamics we use a modified version $V'$ in
%which the values of $\Phi$ have been scaled down by a
%parameter $\lambda_1 \in [0,1)$.
%\[
%v'_i = H_{i,\Sigma}(v_i^0 + \lambda_1 \phi)
%\]
%When drift does occur, this will cause the true
%vertex positions
%$V$ to miss the target slightly, but the following shape
%optimization will correct for this and pull the mesh
%closer to the joints.  The second drift prevention
%mechanism is a small penalization of the magnitude of
%the $\Phi$ values when computing the $\Phi$ values and
%is discussed in the next section.

%%%%%%%%%%%%%%%%%%%%%%%%%%%%%%%%%%%%%%%%%%%%%%%%%%%%%%%%%%%%%%%%%%
\subsection{Shape Optimization}
%%%%%%%%%%%%%%%%%%%%%%%%%%%%%%%%%%%%%%%%%%%%%%%%%%%%%%%%%%%%%%%%%%
\label{subsec:shape_optimization}

Once the pose has been fit, we update the shape deformation
parameters $\Phi$.  These consist of a vector $\phi_i \in \mathbb{R}^3$ for
each vertex.  The shape optimization uses the residual
from Equation \ref{eq:residual} but incorporates additional
regularization terms.

%\begin{equation}
%\label{eq:r_warp}
%\hat{r}_{i} = r_{i} +
%\sum_{i=1}^{|V|}
%\lambda_{\phi}  ||\phi_i||_2^2 +
%\lambda_{\mathcal{N}} \sum_{n \in \mathcal{N}(i)} ||\phi_i - \phi_n||^2_2
%\end{equation}

\begin{equation}
\label{eq:r_warp}
\hat{r}_{i} = r_{i} +
\lambda_{\phi}  ||\phi_i||_2^2 +
\lambda_{\mathcal{N}} \sum_{n \in \mathcal{N}(i)} ||\phi_i - \phi_n||^2_2
\end{equation}

The term weighted by $\lambda_{\phi}$ penalizes
magnitudes of the $\Phi$ vectors and
helps prevent the mesh from drifting off the skeleton.
The term weighted by $\lambda_{\mathcal{N}}$
penalizes the difference between
each $\phi_i$ and those of a set $\mathcal{N}(i)$ of neighboring vertices
in the default mesh.  We typically use the four closest vertices as the
neighborhood. This helps prevent surface discontinuities
and creases.

As before we compute the derivative of this residual for each
vertex $v_i$ with respect to the elements of $\Phi$,
but make one important simplifying approximation.
Technically the neighborhood smoothing term introduces
interdependence between each
$\phi_i$ and its neighbors, but in order to simplify the
computation we treat each $\phi_i$ as if it were independent.
This means that instead of solving one large but
sparse $3|V|$ by $3|V|$ linear system
$(\partial \hat{r} / \partial \Phi)$ we break it up into a separate
3 by 3 linear system for each vertex $(\partial \hat{r}_i / \partial \phi_i)$
and solve them in parallel.
%Note that each
%term in $r_{warp}$ from Equation \ref{eq:r_warp}
%corresponds to a sum over all vertices.
%Therefore when differentiating the residual with respect to $\phi_i$
%we can enforce independence by substituting $r_{warp}$ with
%the element of the sum corresponding to that vertex.
%\[
%r_{warp_i} = r_{data} + \lambda_{\phi} ||\phi_i||^2_2 +
%\lambda_{\mathcal{N}} \sum_{n \in \mathcal{N}(i)} ||
%\phi_i - \phi_n||^2_2~,
%\]
This means we can compute a Jacobian $J_i$ for each vertex as
\[
J_i = \frac{\partial \hat{r}_{i}}{\partial \phi_i} =
\frac{\partial r_{i}}{\phi_i} +
2\lambda_{\phi} \phi_i +
2\lambda_{\mathcal{N}} \sum_{n\in \mathcal{N}(i)}(\phi_i - \phi_n)
\]

Once we have computed our warp Jacobian
$J_i$ we compute $J_i^TJ_i$ as before and solve
\[
(J_i^T J_i + \lambda_{w} \text{diag}(J_i^T J_i))x= J_i^T \hat{r}_{i}
\]
for $x$, and subtract it from $\phi_i$.
This means we have a linear system of
three equations with three unknowns for each vertex, which we
solve in batch on a GPU, assigning one linear system to each thread.
The bottom left frame of Figure \ref{fig:one} shows the result
of the shape deformation after the kinematic pose has been fit.
In practice only two iterations of shape refinement are necessary
for each incoming video frame.

\begin{figure}[b]
\centering
\includegraphics[width=0.4\textwidth]{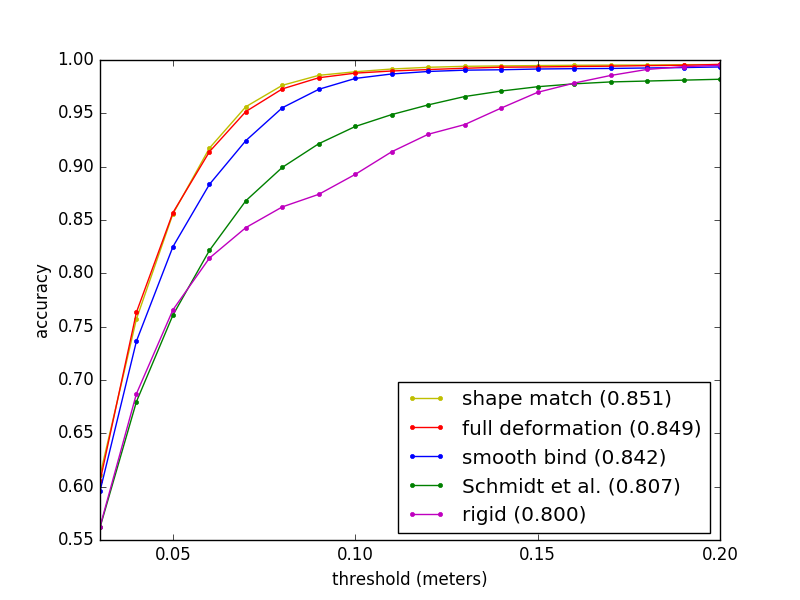}
\caption{The performance of our method in various conditions
on our dataset along with the
method of Schmidt et al. \cite{schmidt2014dart} for comparison.
Each curve shows the percentage of joint positions in these
sequences that are within the distance indicated on the x-axis
from the ground truth positions.
Area under the curve is shown for each plot in parentheses.}
\label{fig:pose_curves}
\end{figure}

%%%%%%%%%%%%%%%%%%%%%%%%%%%%%%%%%%%%%%%%%%%%%%%%%%%%%%%%%%%%%%%%%%
%%%%%%%%%%%%%%%%%%%%%%%%%%%%%%%%%%%%%%%%%%%%%%%%%%%%%%%%%%%%%%%%%%
\section{Experiments}
%%%%%%%%%%%%%%%%%%%%%%%%%%%%%%%%%%%%%%%%%%%%%%%%%%%%%%%%%%%%%%%%%%
%%%%%%%%%%%%%%%%%%%%%%%%%%%%%%%%%%%%%%%%%%%%%%%%%%%%%%%%%%%%%%%%%%
\label{sec:experiments}

\begin{figure}[b]
\centering
\includegraphics[width=0.4\textwidth]{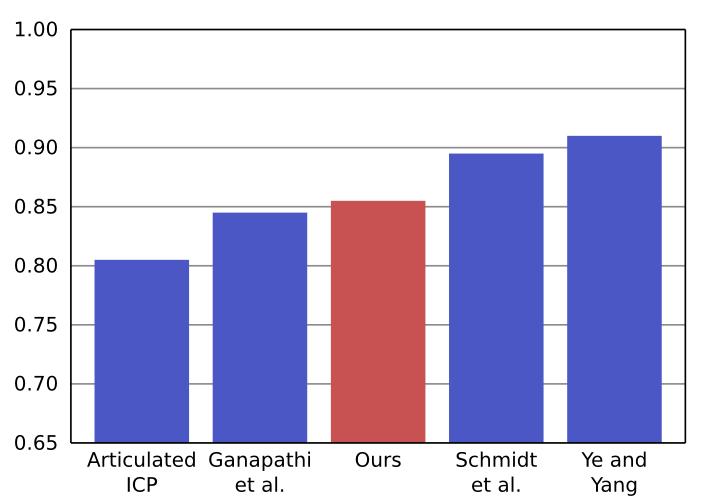}
\caption{Our performance on the EVAL dataset compared to
Articulated ICP reported by \cite{ganapathi2012real},
Ganapthi et al.\cite{ganapathi2012real},
Schmidt et al.\cite{schmidt2014dart} and
Ye and Yang\cite{ye2014real}.}
\label{fig:eval}
\end{figure}

%There are a number of existing datasets designed to test the
%capabilities of markerless motion capture systems.
%The SMMC \cite{ganapathi2010real} and EVAL
%\cite{ganapathi2012real} datasets provide depth images along with
%ground truth pose data.
%The Personalized Depth Tracker (PDT) dataset
%\cite{helten2013personalization} contains ground
%truth information for adapting a mesh shape to different
%subjects, but is not concerned with estimating shape dynamics
%over time.
%Unfortunately the depth information in these datasets was
%captured with either a first generation Microsoft Kinect
%in the case of PDT and EVAL or a Swiss Ranger SR4000
%in the case of SMMC and does not have enough fidelity to capture
%high resolution surface details.  Specifically in the PDT and EVAL
%datasets the depth values are discretized to around two centimeter
%intervals, while the SMMC data is only 176x144 pixels with heavy
%depth noise.  We therefore evaluate our pose tracking on the EVAL
%data without dynamic shape warping and do further experiments
%with our own data to quantify the performance of our
%mesh tracking.

There are a number of existing datasets designed to test the
capabilities of markerless motion capture systems on point cloud data.
The SMMC \cite{ganapathi2010real} and EVAL
\cite{ganapathi2012real} datasets provide depth images along with
ground truth pose data.
The Personalized Depth Tracker (PDT) dataset
\cite{helten2013personalization} contains ground
truth information for adapting a mesh shape to different
subjects, but is not concerned with estimating shape dynamics
over time.
Unfortunately the depth information in these datasets was
captured with either a first generation Microsoft Kinect
in the case of PDT and EVAL or a Swiss Ranger SR4000
in the case of SMMC and does not have enough fidelity to capture
high resolution surface details.  Specifically in the PDT and EVAL
datasets the depth values are discretized to around two centimeter
intervals, while the SMMC data is only 176x144 pixels with heavy
depth noise.  To address this we generated a new dataset of
videos captured with the second generation Microsoft Kinect
(Kinect One) \cite{sell2014xbox} camera using the open source
libfreenect2 drivers \cite{lingzhu_xiang_2016_50641},
but also test on the EVAL dataset for completeness.
Finally we also report reconstruction error for our method to
quantify improvement over conditions that do not estimate dynamic
shape.
%We used the open source
%libfreenect2 drivers \cite{lingzhu_xiang_2016_50641} to stream data
%from this device, and used the factory calibrated camera parameter.

The experiments in this paper were performed on a PC running the
Ubuntu Linux distribution with a 2.4 GHz Xeon quadcore
processor and an Nvidia GeForce 1080 and on a laptop with a 2.6 Ghz
Intel i7 and an NVidia Geforce 1070. Both of these machines run our
tracker at frame rates faster than 20Hz.

%%%%%%%%%%%%%%%%%%%%%%%%%%%%%%%%%%%%%%%%
\begin{figure}[t]
\centering
\includegraphics[width=0.4\textwidth]{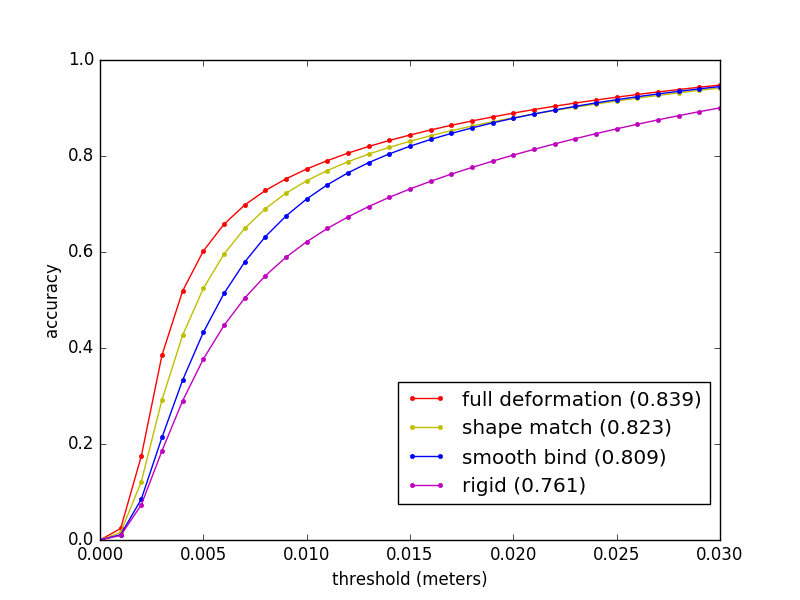}
\caption{
Reconstruction error of our method under various conditions.
Area under the curve is listed for each plot in parentheses.
\label{fig:shape_plot}}
\end{figure}
%%%%%%%%%%%%%%%%%%%%%%%%%%%%%%%%%%%%%%%%

%%%%%%%%%%%%%%%%%%%%%%%%%%%%%%%%%%%%%%%%
\begin{figure}[b]
\centering
\includegraphics[width=0.48\textwidth]{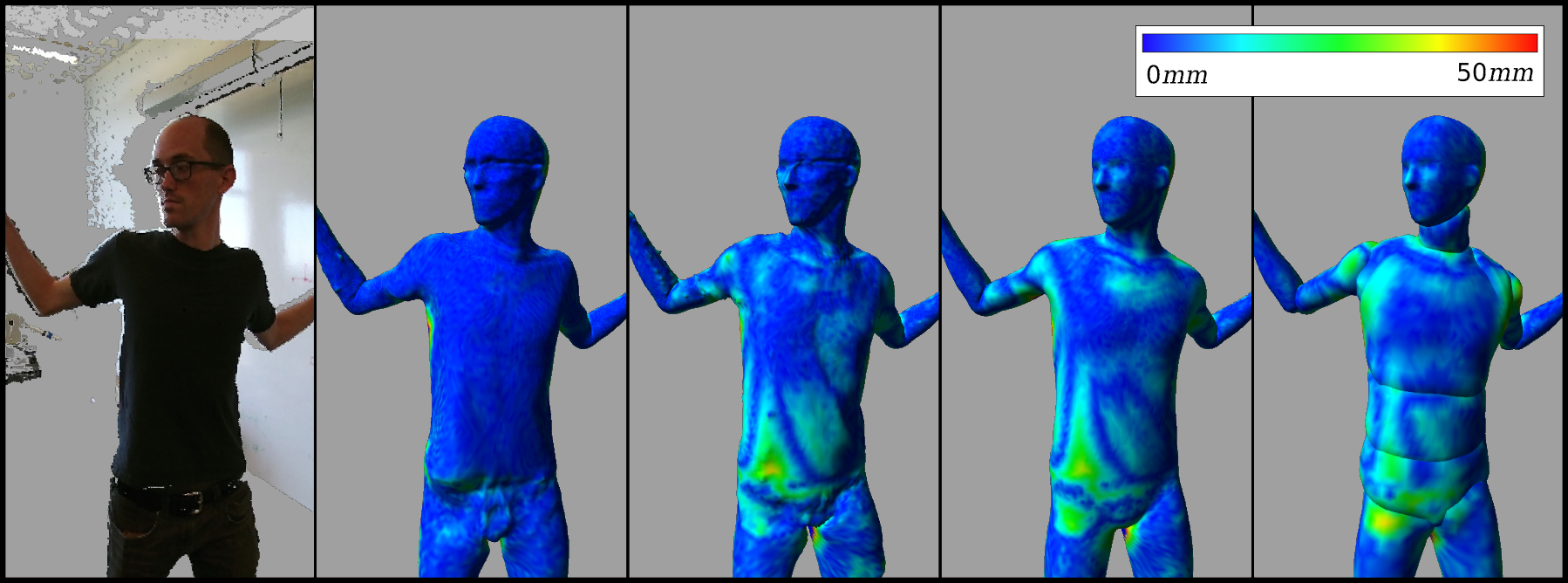}
\caption{
Reconstruction error visualized for a single frame.
From left to right: the point cloud, our fully deformed
model, our model with $\Phi$ fixed to the best fit at
the start frame, the smooth skinned mesh with $\Phi = 0$ and
rigid link geometry.
}
\label{fig:residual_color}
\end{figure}
%%%%%%%%%%%%%%%%%%%%%%%%%%%%%%%%%%%%%%%%

%%%%%%%%%%%%%%%%%%%%%%%%%%%%%%%%%%%%%%%%
\begin{figure*}[t]
\centering
\includegraphics[width=0.99\textwidth]{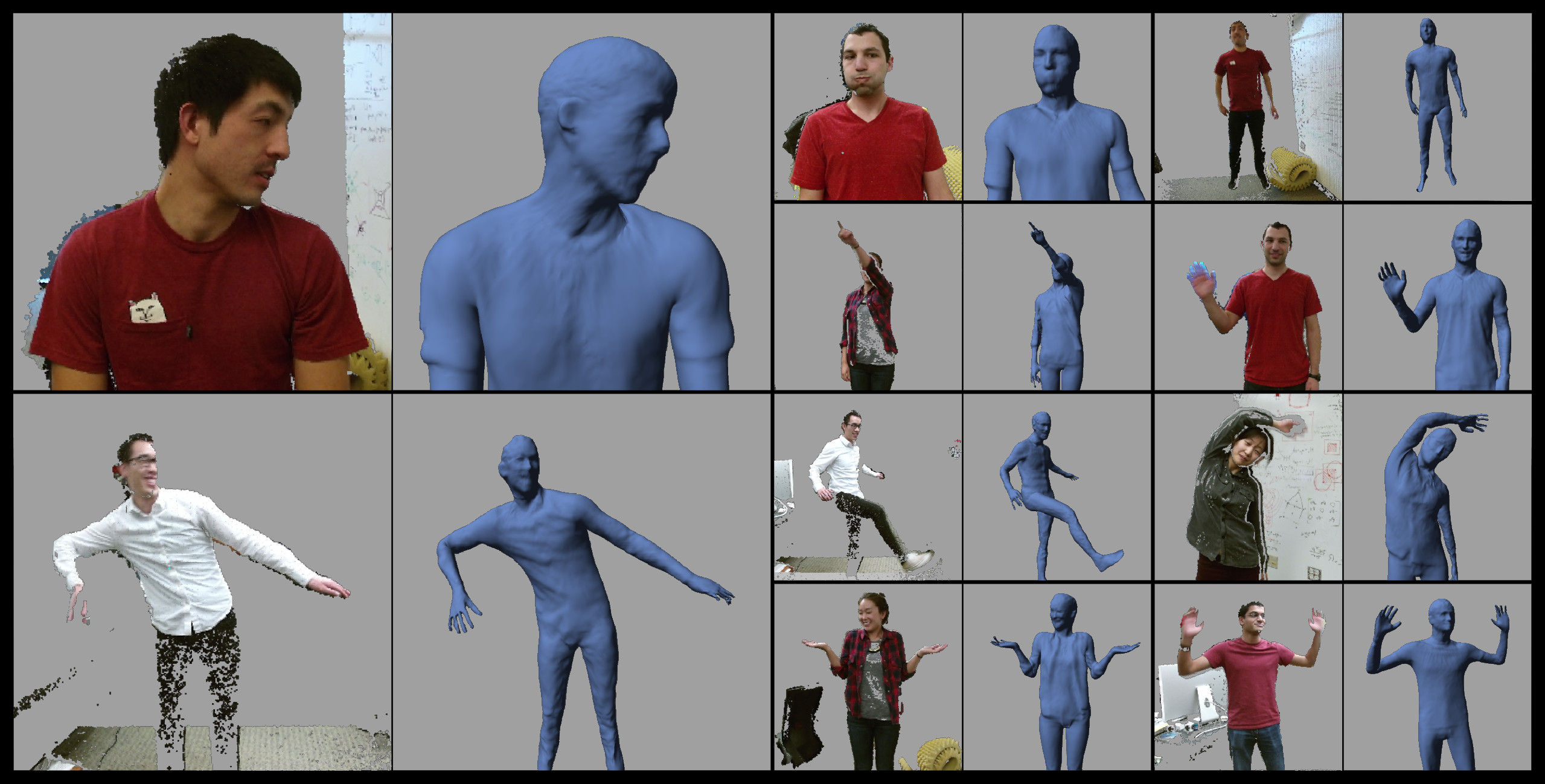}
\caption{A collection of still frames showing the results of
our system.  For each pair of images, the left shows the colored
point cloud while the right shows our warped output mesh.}
\label{fig:results}
\end{figure*}
%%%%%%%%%%%%%%%%%%%%%%%%%%%%%%%%%%%%%%%%

\subsection{Our Dataset}
Our dataset consists of four sequences with
high quality manually annotated pose information.  There are two
subjects, each of which has one close sequence from the waist up and
one far sequence where the full body is visible.
Each sequence contains 300 frames of depth
video with a resolution of 512x424 pixels.  We label the 3D location
of fifteen joints in each frame: the head, torso, pelvis and
left/right hip, knee, ankle, shoulder, elbow and wrist.  We also
annotate when these joints become invisible by either leaving the
frame or becoming occluded.
We tested our system on
these sequences in different conditions to study how
different components of our system affect the performance
of pose tracking.  This includes our full system which estimates
dynamic shape on each frame, a shape match mode that estimates
shape on the first frame and then freezes the dynamic warp
parameters $\Phi$ for the rest of the sequence, a smooth bind mode
which does not perform shape estimation at all and only tracks
kinematic pose and a separate model made up of
rigid mesh segments for each link.
We also tested the method of Schmidt et al.
\cite{schmidt2014dart} on these sequences using an open
source implementation provided by the authors.  We used our own
model with this method for consistency, but had to remove
the prismatic joints because they are unsupported.  In all other
cases, the kinematic hierarchy and joint positions were the same.
The rigid segments used when testing Schmidt et al. and our own
rigid model were generated by cutting our mesh into multiple
disjoint components and filling the resulting holes.

Rather than
reporting a precision score using a single threshold to determine
correctness as is common practice on the EVAL dataset,
we instead plot accuracy as a function of this threshold.
%This is similar to the PCKh measure
%introduced in \cite{andriluka20142d}, but rather
%than using head length as a measuring unit, we use
%metric distance because we are tracking in 3D.
Figure
\ref{fig:pose_curves} shows these results.
As can be seen, the dynamic shape estimation and the mode that
fits the shape to the first frame perform almost identically,
and significantly improve
tracking performance compared to other methods.  This demonstrates
the importance of shape accuracy for template based tracking.

%%%%%%%%%%%%%%%%%%%%%%%%%%%%%%%%%%%%%%%%%%%%%%%%%%%%%%%%%%%%%%%%%%
\subsection{EVAL Dataset}
%%%%%%%%%%%%%%%%%%%%%%%%%%%%%%%%%%%%%%%%%%%%%%%%%%%%%%%%%%%%%%%%%%
%In order to validate the effectiveness of our skeletal tracking
%techniques, we also evaluate our tracker on the Stanford
%EVAL \cite{EVAL}
%dataset.

The EVAL dataset consists of twenty-four RGBD sequences split
evenly across three human subjects with varying body proportions.
%The dataset contains ground truth joint positions for twelve
%body parts in each video frame captured using a commercial
%motion capture system.
The evaluation criteria is the percentage
of frames in which the estimated joint position is within ten
centimeters of the ground truth.  Because the ground truth
data relies on joint locations specific to a particular model, we
follow the technique of \cite{ye2014real} and use mean-subtraction to
find the best placement of the tracked joints relative to our model.
%This data was recorded
%with a first generation Microsoft Kinect,
%and so it does not have the same fidelity as our new data.  Specifically
%the depth image is 320 x 240 and more importantly the depth values
%are discretized at intervals of almost two centimeters.
%This severely
%limits the utility of our dynamic deformation and so we disabled it
%for this experiment.
Because of the limitations of the depth data pointed out above we
disabled the dynamic shape estimation and used a simplified
kinematics model and mesh for this experiment.
%Given that our system was not designed for low
%resolution data, we were not able to achieve state of the art
%performance on this dataset.

Figure \ref{fig:eval} shows our performance compared to
the reported scores of other methods.
While we do not perform as well as other state of the art techniques,
our method was not designed to work with low resolution depth data.
%and our results are comparable with the range of existing literature.

%%%%%%%%%%%%%%%%%%%%%%%%%%%%%%%%%%%%%%%%%%%%%%%%%%%%%%%%%%%%%%%%%%
\subsection{Shape Fitting}
%%%%%%%%%%%%%%%%%%%%%%%%%%%%%%%%%%%%%%%%%%%%%%%%%%%%%%%%%%%%%%%%%%

%In order to test our shape estimation we generated
%a new dataset consisting of sixteeen high resolution RGBD sequences.
%These sequences consist of four action sets performed by
%four different subjects.
%Unlike the SMMC and EVAL datasets that feature fast
%acrobatic motion designed to
%test the limits of pose estimation techniques, these
%sequences feature real-world motions.
%The first of these sequences was captured at a close range
%and features basic head and facial motion.
%The second and third were captured from an intermediate distance
%and feature basic upper body motion such as shrugs,
%waves, handovers and pointing.  The fourth was captured at a
%farther distance and contains jumping and stretches.
%Figure \ref{fig:distances} shows three frames from these sequences
%to show the varying ranges at which these sequences were captured.

%We evaluate our shape estimation by tracking our model through
%sequence twice.  For one run the track is run normally but for the
%other the warp is disabled so that the model maintains its default
%shape throughout the sequence.
In order to test our dynamic shape estimation, we compute the
reconstruction error of our model as the
point wise distances between the visible vertices
and the nearest point in the observation for each frame in our dataset.
We test this under the same conditions that we used to test our
pose tracking.  Figure \ref{fig:shape_plot} shows these results.
In this case our dynamic shape fitting
offers clear improvements over the 
other modes.  Figure \ref{fig:residual_color} shows a single frame
from each mode colored to show reconstruction error.
%For each run we record the distance from each
%visible model vertex to its closest observation point in each frame.
%Because we do not have dense ground truth labels for this data, these
%results do not show overall tracking performance.  Instead this data
%shows how much the dynamic shape estimation improves fitting
%the point-cloud observations.
%Figure \ref{fig:shape_plot} shows these errors binned into
%histograms while Figure \ref{fig:residual_color} shows one
%frame with both meshes colored to indicate the fitting error.

%The shape estimation shows
%improvement in the form of higher concentration of error at
%low values for all sequences.  However this improvement is
%most pronounced at closer ranges where more detail is visible.

%%%%%%%%%%%%%%%%%%%%%%%%%%%%%%%%%%%%%%%%%%%%%%%%%%%%%%%%%%%%%%%%%%
%\subsection{Multi Resolution}
%%%%%%%%%%%%%%%%%%%%%%%%%%%%%%%%%%%%%%%%%%%%%%%%%%%%%%%%%%%%%%%%%%
%Show with and without the multi-resolution interpolation.

%%%%%%%%%%%%%%%%%%%%%%%%%%%%%%%%%%%%%%%%%%%%%%%%%%%%%%%%%%%%%%%%%%
\subsection{Qualitative Results}
%%%%%%%%%%%%%%%%%%%%%%%%%%%%%%%%%%%%%%%%%%%%%%%%%%%%%%%%%%%%%%%%%%
Figure \ref{fig:results} shows ten frames from four male and two
female subjects.  Each frame has a colored point clouds and
the corresponding tracked mesh.  In all cases the tracker
was given a rough initialization
of the subject's pose at the start of the sequence, and the
pose and shape tracked from that point forward.  The model was
not customized to any of these subjects beforehand except for a
single uniform scale parameter which was only approximately
estimated based on the subject's height.  This
demonstrates the robustness of our model to different of
subjects with varying proportions at a range of distances to
camera.  Detailed surface features
are clearly visible in all images showing the visual fidelity
of our tracked meshes.

%Semantic consistency with figure?

%%%%%%%%%%%%%%%%%%%%%%%%%%%%%%%%%%%%%%%%%%%%%%%%%%%%%%%%%%%%%%%%%%
%\subsection{Sensors and Hardware}
%%%%%%%%%%%%%%%%%%%%%%%%%%%%%%%%%%%%%%%%%%%%%%%%%%%%%%%%%%%%%%%%%%

%%%%%%%%%%%%%%%%%%%%%%%%%%%%%%%%%%%%%%%%%%%%%%%%%%%%%%%%%%%%%%%%%%
%%%%%%%%%%%%%%%%%%%%%%%%%%%%%%%%%%%%%%%%%%%%%%%%%%%%%%%%%%%%%%%%%%
\section{Conclusion}
\label{sec:conclusion}
%%%%%%%%%%%%%%%%%%%%%%%%%%%%%%%%%%%%%%%%%%%%%%%%%%%%%%%%%%%%%%%%%%
%%%%%%%%%%%%%%%%%%%%%%%%%%%%%%%%%%%%%%%%%%%%%%%%%%%%%%%%%%%%%%%%%%

%Robots must have accurate spatial information to physically interact
%with deformable objects.  We have demonstrated an approach for tracking
%these objects using model-based optimization and shown that it
%can produce dense and accurate estimation of  detailed deformable
%surfaces in real time.  This system also provides useful
%pose estimates of the model's kinematic structure for gesture
%recognition and motion prediction.

We have demonstrated an articulated tracking approach for
deformable objects that is able to track humans using
a high resolution template mesh.  We have shown that it can
produce dense and accurate estimation of detailed deformable
surfaces in real time.  This system also provides useful
pose estimates of the model's kinematic structure for gesture
recognition and motion prediction.

This work opens up some important areas of future research.
While many figures in this paper feature colored point clouds,
our technique does not currently use color
information as part of the tracking process.  Incorporating color
may help the mesh lock on to specific color features and prevent
vertices from drifting along the surface of an object.
Fine structures such as fingers with many
degrees of freedom currently pose a challenge.
This is partially due to sensor resolution, but more could be
done to regularize their kinematic motion.
Taylor er al. \cite{taylor2016efficient} show promising results
in this direction.
Finally this technique can be combined with discriminative detection
systems to more easily recover from tracking errors and avoid
the need for pose initialization.

%%%%%%%%%%%%%%%%%%%%%%%%%%%%%%%%%%%%%%%%%%%%%%%%%%%%%%%%%%%%%%%%%%
%%%%%%%%%%%%%%%%%%%%%%%%%%%%%%%%%%%%%%%%%%%%%%%%%%%%%%%%%%%%%%%%%%
\section*{Acknowledgments}
%%%%%%%%%%%%%%%%%%%%%%%%%%%%%%%%%%%%%%%%%%%%%%%%%%%%%%%%%%%%%%%%%%
%%%%%%%%%%%%%%%%%%%%%%%%%%%%%%%%%%%%%%%%%%%%%%%%%%%%%%%%%%%%%%%%%%

This material is based upon work supported by the
National Science Foundation under Grant No. IIS-1538618 AM002.

%% Use plainnat to work nicely with natbib. 

\bibliographystyle{plainnat}
\bibliography{references}

\begin{thebibliography}{40}
\providecommand{\natexlab}[1]{#1}
\providecommand{\url}[1]{\texttt{#1}}
\expandafter\ifx\csname urlstyle\endcsname\relax
  \providecommand{\doi}[1]{doi: #1}\else
  \providecommand{\doi}{doi: \begingroup \urlstyle{rm}\Url}\fi

\bibitem[{Autodesk Incorporated}()]{maya2016}
{Autodesk Incorporated}.
\newblock Maya.
\newblock URL \url{https://autodesk.com/products/maya/overview}.

\bibitem[Cao et~al.(2016)Cao, Simon, Wei, and Sheikh]{Cao2016RealtimeM2}
Zhe Cao, Tomas Simon, Shih-En Wei, and Yaser Sheikh.
\newblock Realtime multi-person 2d pose estimation using part affinity fields.
\newblock \emph{CoRR}, abs/1611.08050, 2016.

\bibitem[Carreira et~al.(2016)Carreira, Agrawal, Fragkiadaki, and
  Malik]{carreira2016human}
Joao Carreira, Pulkit Agrawal, Katerina Fragkiadaki, and Jitendra Malik.
\newblock Human pose estimation with iterative error feedback.
\newblock In \emph{Proceedings of the IEEE Conference on Computer Vision and
  Pattern Recognition}, pages 4733--4742, 2016.

\bibitem[Catmull and Clark(1978)]{catmull1978recursively}
Edwin Catmull and James Clark.
\newblock Recursively generated b-spline surfaces on arbitrary topological
  meshes.
\newblock \emph{Computer-aided design}, 10\penalty0 (6):\penalty0 350--355,
  1978.

\bibitem[{CG Trader}()]{cgtrader}
{CG Trader}.
\newblock Cg trader.
\newblock URL \url{http://www.cgtrader.com/}.

\bibitem[Chen and Medioni(1992)]{chen1992object}
Yang Chen and G{\'e}rard Medioni.
\newblock Object modelling by registration of multiple range images.
\newblock \emph{Image and vision computing}, 10\penalty0 (3):\penalty0
  145--155, 1992.

\bibitem[Clifford(1882)]{clifford1882mathematical}
William~Kingdon Clifford.
\newblock \emph{Mathematical papers}.
\newblock Macmillan and Company, 1882.

\bibitem[Curless and Levoy(1996)]{curless1996volumetric}
Brian Curless and Marc Levoy.
\newblock A volumetric method for building complex models from range images.
\newblock In \emph{Proceedings of the 23rd annual conference on Computer
  graphics and interactive techniques}, pages 303--312. ACM, 1996.

\bibitem[Daniilidis(1999)]{daniilidis1999hand}
Konstantinos Daniilidis.
\newblock Hand-eye calibration using dual quaternions.
\newblock \emph{The International Journal of Robotics Research}, 18\penalty0
  (3):\penalty0 286--298, 1999.

\bibitem[De~Aguiar et~al.(2008)De~Aguiar, Stoll, Theobalt, Ahmed, Seidel, and
  Thrun]{de2008performance}
Edilson De~Aguiar, Carsten Stoll, Christian Theobalt, Naveed Ahmed, Hans-Peter
  Seidel, and Sebastian Thrun.
\newblock Performance capture from sparse multi-view video.
\newblock In \emph{ACM Transactions on Graphics (TOG)}, volume~27, page~98.
  ACM, 2008.

\bibitem[Dou et~al.(2016)Dou, Khamis, Degtyarev, Davidson, Fanello, Kowdle,
  Escolano, Rhemann, Kim, Taylor, et~al.]{dou2016fusion4d}
Mingsong Dou, Sameh Khamis, Yury Degtyarev, Philip Davidson, Sean~Ryan Fanello,
  Adarsh Kowdle, Sergio~Orts Escolano, Christoph Rhemann, David Kim, Jonathan
  Taylor, et~al.
\newblock Fusion4d: Real-time performance capture of challenging scenes.
\newblock \emph{ACM Transactions on Graphics (TOG)}, 35\penalty0 (4):\penalty0
  114, 2016.

\bibitem[Ferrari et~al.(2008)Ferrari, Marin-Jimenez, and
  Zisserman]{ferrari2008progressive}
Vittorio Ferrari, Manuel Marin-Jimenez, and Andrew Zisserman.
\newblock Progressive search space reduction for human pose estimation.
\newblock In \emph{Computer Vision and Pattern Recognition, 2008. CVPR 2008.
  IEEE Conference on}, pages 1--8. IEEE, 2008.

\bibitem[Gall et~al.(2009)Gall, Stoll, De~Aguiar, Theobalt, Rosenhahn, and
  Seidel]{gall2009motion}
Juergen Gall, Carsten Stoll, Edilson De~Aguiar, Christian Theobalt, Bodo
  Rosenhahn, and Hans-Peter Seidel.
\newblock Motion capture using joint skeleton tracking and surface estimation.
\newblock In \emph{Computer Vision and Pattern Recognition, 2009. CVPR 2009.
  IEEE Conference on}, pages 1746--1753. IEEE, 2009.

\bibitem[Ganapathi et~al.(2010)Ganapathi, Plagemann, Koller, and
  Thrun]{ganapathi2010real}
Varun Ganapathi, Christian Plagemann, Daphne Koller, and Sebastian Thrun.
\newblock Real time motion capture using a single time-of-flight camera.
\newblock In \emph{Computer Vision and Pattern Recognition (CVPR), 2010 IEEE
  Conference on}, pages 755--762. IEEE, 2010.

\bibitem[Ganapathi et~al.(2012)Ganapathi, Plagemann, Koller, and
  Thrun]{ganapathi2012real}
Varun Ganapathi, Christian Plagemann, Daphne Koller, and Sebastian Thrun.
\newblock Real-time human pose tracking from range data.
\newblock In \emph{European conference on computer vision}, pages 738--751.
  Springer, 2012.

\bibitem[Garcia~Cifuentes et~al.(2016)Garcia~Cifuentes, Issac, W{\"u}thrich,
  Schaal, and Bohg]{GarciaCifuentes.RAL}
Cristina Garcia~Cifuentes, Jan Issac, Manuel W{\"u}thrich, Stefan Schaal, and
  Jeannette Bohg.
\newblock Probabilistic articulated real-time tracking for robot manipulation.
\newblock \emph{IEEE Robotics and Automation Letters (RA-L)}, 2016.

\bibitem[Gill and Murray(1974)]{gill1974newton}
Philip~E Gill and Walter Murray.
\newblock Newton-type methods for unconstrained and linearly constrained
  optimization.
\newblock \emph{Mathematical Programming}, 7\penalty0 (1):\penalty0 311--350,
  1974.

\bibitem[Grest et~al.(2005)Grest, Woetzel, and Koch]{grest2005nonlinear}
Daniel Grest, Jan Woetzel, and Reinhard Koch.
\newblock Nonlinear body pose estimation from depth images.
\newblock In \emph{Joint Pattern Recognition Symposium}, pages 285--292.
  Springer, 2005.

\bibitem[Haque et~al.(2016)Haque, Peng, Luo, Alahi, Yeung, and
  Fei-Fei]{haque2016towards}
Albert Haque, Boya Peng, Zelun Luo, Alexandre Alahi, Serena Yeung, and
  Li~Fei-Fei.
\newblock Towards viewpoint invariant 3d human pose estimation.
\newblock In \emph{European Conference on Computer Vision}, pages 160--177.
  Springer, 2016.

\bibitem[Helten et~al.(2013)Helten, Baak, Bharaj, M{\"u}ller, Seidel, and
  Theobalt]{helten2013personalization}
Thomas Helten, Andreas Baak, Gaurav Bharaj, Meinard M{\"u}ller, Hans-Peter
  Seidel, and Christian Theobalt.
\newblock Personalization and evaluation of a real-time depth-based full body
  tracker.
\newblock In \emph{2013 International Conference on 3D Vision-3DV 2013}, pages
  279--286. IEEE, 2013.

\bibitem[Huang et~al.(2015)Huang, Boyer, do~Canto~Angonese, Navab, and
  Ilic]{huang2015toward}
Chun-Hao Huang, Edmond Boyer, Bibiana do~Canto~Angonese, Nassir Navab, and
  Slobodan Ilic.
\newblock Toward user-specific tracking by detection of human shapes in
  multi-cameras.
\newblock In \emph{Proceedings of the IEEE Conference on Computer Vision and
  Pattern Recognition}, pages 4027--4035, 2015.

\bibitem[Huang et~al.(2016)Huang, Allain, Franco, Navab, Ilic, and
  Boyer]{huang2016volumetric}
Chun-Hao Huang, Benjamin Allain, Jean-S{\'e}bastien Franco, Nassir Navab,
  Slobodan Ilic, and Edmond Boyer.
\newblock Volumetric 3d tracking by detection.
\newblock In \emph{Proceedings of the IEEE Conference on Computer Vision and
  Pattern Recognition}, pages 3862--3870, 2016.

\bibitem[Innmann et~al.(2016)Innmann, Zollh{\"o}fer, Nie{\ss}ner, Theobalt, and
  Stamminger]{innmann2016volume}
Matthias Innmann, Michael Zollh{\"o}fer, Matthias Nie{\ss}ner, Christian
  Theobalt, and Marc Stamminger.
\newblock Volumedeform: Real-time volumetric non-rigid reconstruction.
\newblock In \emph{Proceedings of the European Conference on Computer Vision
  ({ECCV})}, 2016.

\bibitem[Iqbal et~al.(2016)Iqbal, Milan, and Gall]{Iqbal2016PoseTrackJM}
Umar Iqbal, Anton Milan, and Juergen Gall.
\newblock Pose-track: Joint multi-person pose estimation and tracking.
\newblock \emph{CoRR}, abs/1611.07727, 2016.

\bibitem[Kavan et~al.(2008)Kavan, Collins, {\v{Z}}{\'a}ra, and
  O'Sullivan]{kavan2008geometric}
Ladislav Kavan, Steven Collins, Ji{\v{r}}{\'\i} {\v{Z}}{\'a}ra, and Carol
  O'Sullivan.
\newblock Geometric skinning with approximate dual quaternion blending.
\newblock \emph{ACM Transactions on Graphics (TOG)}, 27\penalty0 (4):\penalty0
  105, 2008.

\bibitem[Li et~al.(2012)Li, Luo, Vlasic, Peers, Popovi{\'c}, Pauly, and
  Rusinkiewicz]{li2012temporally}
Hao Li, Linjie Luo, Daniel Vlasic, Pieter Peers, Jovan Popovi{\'c}, Mark Pauly,
  and Szymon Rusinkiewicz.
\newblock Temporally coherent completion of dynamic shapes.
\newblock \emph{ACM Transactions on Graphics (TOG)}, 31\penalty0 (1):\penalty0
  2, 2012.

\bibitem[Lin et~al.(2014)Lin, Maire, Belongie, Hays, Perona, Ramanan,
  Doll{\'a}r, and Zitnick]{lin2014microsoft}
Tsung-Yi Lin, Michael Maire, Serge Belongie, James Hays, Pietro Perona, Deva
  Ramanan, Piotr Doll{\'a}r, and C~Lawrence Zitnick.
\newblock Microsoft coco: Common objects in context.
\newblock In \emph{European conference on computer vision}, pages 740--755.
  Springer, 2014.

\bibitem[Marquardt(1963)]{marquardt1963algorithm}
Donald~W Marquardt.
\newblock An algorithm for least-squares estimation of nonlinear parameters.
\newblock \emph{Journal of the society for Industrial and Applied Mathematics},
  11\penalty0 (2):\penalty0 431--441, 1963.

\bibitem[Mehta et~al.(2017)Mehta, Sridhar, Sotnychenko, Rhodin, Shafiei,
  Seidel, Xu, Casas, and Theobalt]{mehta2017vnect}
Dushyant Mehta, Srinath Sridhar, Oleksandr Sotnychenko, Helge Rhodin, Mohammad
  Shafiei, Hans-Peter Seidel, Weipeng Xu, Dan Casas, and Christian Theobalt.
\newblock Vnect: Real-time 3d human pose estimation with a single rgb camera.
\newblock \emph{arXiv preprint arXiv:1705.01583}, 2017.

\bibitem[Newcombe et~al.(2015)Newcombe, Fox, and
  Seitz]{newcombe2015dynamicfusion}
Richard~A Newcombe, Dieter Fox, and Steven~M Seitz.
\newblock Dynamicfusion: Reconstruction and tracking of non-rigid scenes in
  real-time.
\newblock In \emph{Proceedings of the IEEE conference on computer vision and
  pattern recognition}, pages 343--352, 2015.

\bibitem[{NoneCG}()]{nonecg}
{NoneCG}.
\newblock Nonecg.
\newblock URL \url{http://www.nonecg.com/}.

\bibitem[Plagemann et~al.(2010)Plagemann, Ganapathi, Koller, and
  Thrun]{plagemann2010real}
Christian Plagemann, Varun Ganapathi, Daphne Koller, and Sebastian Thrun.
\newblock Real-time identification and localization of body parts from depth
  images.
\newblock In \emph{Robotics and Automation (ICRA), 2010 IEEE International
  Conference on}, pages 3108--3113. IEEE, 2010.

\bibitem[Schmidt et~al.(2014)Schmidt, Newcombe, and Fox]{schmidt2014dart}
Tanner Schmidt, Richard Newcombe, and Dieter Fox.
\newblock Dart: Dense articulated real-time tracking.
\newblock \emph{Proceedings of Robotics: Science and Systems, Berkeley, USA},
  2, 2014.

\bibitem[Sell and O'Connor(2014)]{sell2014xbox}
John Sell and Patrick O'Connor.
\newblock The xbox one system on a chip and kinect sensor.
\newblock \emph{IEEE Micro}, 34\penalty0 (2):\penalty0 44--53, 2014.

\bibitem[Shotton et~al.(2013)Shotton, Sharp, Kipman, Fitzgibbon, Finocchio,
  Blake, Cook, and Moore]{shotton2013real}
Jamie Shotton, Toby Sharp, Alex Kipman, Andrew Fitzgibbon, Mark Finocchio,
  Andrew Blake, Mat Cook, and Richard Moore.
\newblock Real-time human pose recognition in parts from single depth images.
\newblock \emph{Communications of the ACM}, 56\penalty0 (1):\penalty0 116--124,
  2013.

\bibitem[Taylor et~al.(2016)Taylor, Bordeaux, Cashman, Corish, Keskin, Sharp,
  Soto, Sweeney, Valentin, Luff, et~al.]{taylor2016efficient}
Jonathan Taylor, Lucas Bordeaux, Thomas Cashman, Bob Corish, Cem Keskin, Toby
  Sharp, Eduardo Soto, David Sweeney, Julien Valentin, Benjamin Luff, et~al.
\newblock Efficient and precise interactive hand tracking through joint,
  continuous optimization of pose and correspondences.
\newblock \emph{ACM Transactions on Graphics (TOG)}, 35\penalty0 (4):\penalty0
  143, 2016.

\bibitem[Tompson et~al.(2014)Tompson, Stein, Lecun, and
  Perlin]{tompson2014real}
Jonathan Tompson, Murphy Stein, Yann Lecun, and Ken Perlin.
\newblock Real-time continuous pose recovery of human hands using convolutional
  networks.
\newblock \emph{ACM Transactions on Graphics (TOG)}, 33\penalty0 (5):\penalty0
  169, 2014.

\bibitem[Xiang et~al.(2016)Xiang, Echtler, Kerl, Wiedemeyer, Lars, hanyazou,
  Gordon, Facioni, laborer2008, Wareham, Goldhoorn, alberth, gaborpapp, Fuchs,
  jmtatsch, Blake, Federico, Jungkurth, Mingze, vinouz, Coleman, Burns, Rawat,
  Mokhov, Reynolds, Viau, Fraissinet-Tachet, Ludique, Billingham, and
  Alistair]{lingzhu_xiang_2016_50641}
Lingzhu Xiang, Florian Echtler, Christian Kerl, Thiemo Wiedemeyer, Lars,
  hanyazou, Ryan Gordon, Francisco Facioni, laborer2008, Rich Wareham, Matthias
  Goldhoorn, alberth, gaborpapp, Steffen Fuchs, jmtatsch, Joshua Blake,
  Federico, Henning Jungkurth, Yuan Mingze, vinouz, Dave Coleman, Brendan
  Burns, Rahul Rawat, Serguei Mokhov, Paul Reynolds, P.E. Viau, Matthieu
  Fraissinet-Tachet, Ludique, James Billingham, and Alistair.
\newblock libfreenect2: Release 0.2, April 2016.
\newblock URL \url{https://doi.org/10.5281/zenodo.50641}.

\bibitem[Ye and Yang(2014)]{ye2014real}
Mao Ye and Ruigang Yang.
\newblock Real-time simultaneous pose and shape estimation for articulated
  objects using a single depth camera.
\newblock In \emph{Proceedings of the IEEE Conference on Computer Vision and
  Pattern Recognition}, pages 2345--2352, 2014.

\bibitem[Zollh{\"o}fer et~al.(2014)Zollh{\"o}fer, Nie{\ss}ner, Izadi, Rehmann,
  Zach, Fisher, Wu, Fitzgibbon, Loop, Theobalt, et~al.]{zollhofer2014real}
Michael Zollh{\"o}fer, Matthias Nie{\ss}ner, Shahram Izadi, Christoph Rehmann,
  Christopher Zach, Matthew Fisher, Chenglei Wu, Andrew Fitzgibbon, Charles
  Loop, Christian Theobalt, et~al.
\newblock Real-time non-rigid reconstruction using an rgb-d camera.
\newblock \emph{ACM Transactions on Graphics (TOG)}, 33\penalty0 (4):\penalty0
  156, 2014.

\end{thebibliography}

\end{document}